\documentclass{article} 
\usepackage[utf8]{inputenc}
\usepackage{iclr2026_conference,times}
\iclrfinalcopy

\usepackage{amsmath,amsfonts,bm}

\def\eqref#1{equation~\ref{#1}}

\def\1{\bm{1}}

\DeclareMathAlphabet{\mathsfit}{\encodingdefault}{\sfdefault}{m}{sl}
\SetMathAlphabet{\mathsfit}{bold}{\encodingdefault}{\sfdefault}{bx}{n}

\usepackage{hyperref}
\usepackage{url}
\usepackage{times}
\usepackage{latexsym}
\usepackage{microtype}
\usepackage[T1]{fontenc}
\usepackage[utf8]{inputenc}
\usepackage{inconsolata}
\usepackage{graphicx}
\usepackage{enumitem}
\usepackage{cleveref}
\usepackage{booktabs}
\usepackage{minitoc}
\usepackage{multirow}

\usepackage{listings}
\newcommand{\name}{{\textbf{OKBench}}}

\title{OKBench:\\Democratizing LLM Evaluation \\with Fully Automated, On-Demand,\\ Open Knowledge Benchmarking}

\author{Yanhong Li\thanks{These authors contributed equally to this work.} \\
  University of Chicago \\
  \texttt{yanhongli@uchicago.edu}\\\And
  Tianyang Xu$^{*}$ \\
  TTI-Chicago \\
  \texttt{sallyxu@ttic.edu} \\\And
  Kenan Tang$^{*}$ \\
  UC Santa Barbara \\
  \texttt{kenantang@ucsb.edu} \\
  \AND
  Karen Livescu \\
  TTI-Chicago \\
  \texttt{klivescu@ttic.edu} \\
  \And
  David McAllester \\
  TTI-Chicago \\
\texttt{mcallester@ttic.edu} \\
  \And
  Jiawei Zhou \\
  Stony Brook University \\
\texttt{jiawei.zhou.1@stonybrook.edu}
  }

\begin{document}
\maketitle

\begin{abstract}

Knowledge-intensive question answering is central to large language models (LLMs) and is typically assessed using static benchmarks derived from sources like Wikipedia and textbooks. However, these benchmarks fail to capture evolving knowledge in a dynamic world, and centralized curation struggles to keep pace with rapid LLM advancements.
To address these drawbacks, we propose Open Knowledge Bench (\name), a fully automated framework for generating high-quality, dynamic knowledge benchmarks on demand. Focusing on the news domain where knowledge updates daily, \name~is an agentic framework that automates the sourcing, creation, validation, and distribution of benchmarks.
Our approach democratizes benchmark creation and facilitates thorough evaluation of retrieval-augmented methods by reducing overlap with pretraining data.
We evaluate our framework on a wide range open-source and proprietary LLMs of various sizes and configurations, both with and without retrieval over freshly generated knowledge.
Our results reveal distinct model behaviors when confronted with new information and highlight how retrieval narrows the performance gap between small and large models.
These findings underscore the importance of evaluating LLMs on evolving knowledge benchmarks.

\end{abstract}

\section{Introduction}

One of the most common uses of large language models (LLMs) is for answering knowledge-intensive questions.  However, this task is challenging as factual knowledge in the real world evolves rapidly. 
Well-trained models can quickly become outdated \citep{li-etal-2024-open-source}, raising the need for continual model updates \citep{liška2022streamingqabenchmarkadaptationnew} or improved retrieval-augmented generation (RAG) techniques \citep{lewis2020retrieval}. 
At the same time, the lack of transparency in training data makes it difficult to assess how recent a model’s knowledge truly is \citep{chengdated2024}. 
Existing benchmarks also struggle to keep pace: once released, their contents may be absorbed into future training data, weakening their utility and leading to benchmark contamination. 
This phenomenon complicates the evaluation of retrieval-based methods, as models may have already memorized the relevant facts during training. 
In this paper, we propose that the solution to these challenges is {\bf fast, automated, decentralized curation of dynamic knowledge benchmarks} that can track LLM development in real time and offer a clean testbed for evaluating retrieval augmented methods.

Despite the rapid advancement of LLMs and the growing need for accurate knowledge assessment, most popular benchmarks remain \emph{static} after creation. 
Widely used datasets such as Natural Questions \citep{kwiatkowski-etal-2019-natural}, TriviaQA \citep{joshi-etal-2017-triviaqa}, and HotpotQA \citep{yang-etal-2018-hotpotqa} are primarily drawn from Wikipedia or curated text snapshots over a fixed time period. 
While instrumental in advancing open-domain question answering (QA) research, these benchmarks quickly become outdated and are often included in model pretraining corpora, leading to data contamination and inflated performance estimation \citep{li-etal-2024-open-source}. 
More recent efforts such as StreamingQA \citep{liška2022streamingqabenchmarkadaptationnew}, RealTimeQA \citep{kasai2024realtimeqawhatsanswer} and FreshQA \citep{vu-etal-2024-freshllms} have begun including fresh facts. 
However, these dynamic benchmarks still rely on partial human curation and infrequent updates, or focus on a different task like forecasting. 
As a result, they don't enable continuous evaluation of LLMs' capabilities.
Finally, these previous efforts are \textit {centralized}, making it difficult and expensive to reproduce them on demand.

\begin{table}[t]
    \caption{Comparison of our benchmark with some previous knowledge QA \& dynamic benchmarks in terms of objective, automation, update frequency, and scale. \\}
    \label{tab:dynamic-benchmarks}
    \centering
    \resizebox{0.9\linewidth}{!}{%
    \begin{tabular}{lllll}
        \toprule
        \textbf{Benchmark} & \textbf{Objective} & \textbf{Automation} & \textbf{Update Freq.} & \textbf{Scale} \\
        \midrule
        StreamingQA & Factual QA & Partial & Static & $36,800$ QA pairs \\
        RealTime QA & Factual QA & Partial & Weekly & $\sim 30$ QA pairs \\
        FreshQA & Factual QA \& Debunking & Low & Weekly & $600$ QA pairs, only update answers \\
        LiveBench & Reasoning & Partial & Monthly & $40-100$ questions per task \\
        Daily Oracle & Forecasting & Full & Daily & $\sim 17.2$ QA pairs \\
        FutureX & Future Event Prediction & Full & Daily \& Weekly & $500$ events \\
        Ours & Factual QA & Full & Any time & $\sim 2000$ QA pairs \\
        \bottomrule
    \end{tabular}%
    }
\end{table}

We propose an approach that addresses these challenges and democratizes dynamic knowledge benchmarking, by making it easy and practical for anyone to generate a new reproducible benchmark anytime. Specifically, we introduce \textbf{{\name}} (\textbf{O}pen \textbf{K}nowledge \textbf{Bench}), a fully automated framework for generating knowledge benchmarks for fair LLM evaluation. 
Focusing on the news domain where new knowledge emerges daily, our system automates the entire pipeline from information extraction to benchmark construction, producing multiple‑choice QA items (with optional open‑ended variants). 
\name~is an agentic framework built on state-of-the-art LLMs, in which specialized agents for QA generation and validation collaborate to promote quality and consistency. 

To enable benchmark generation at any time, we introduce a distribution and version control protocol that assigns each benchmark a unique signature, assuring consistent tracking and fair comparison across models and evaluations. 
The framework is fully \emph{open-source}, empowering \emph{any user} to generate up-to-date benchmarks \emph{at any time}.
This enables diverse use cases such as monitoring LLM knowledge freshness or evaluating retrieval-augmented models on clean, non-memorized data. 

To assess the quality of the automatically generated benchmarks, we conduct manual validation of one of our question sets and find it to be of high quality.

To demonstrate the utility of our framework and assess current model capabilities, we evaluate a range of open-source and proprietary LLMs across multiple model sizes, with and without retrieval augmentation, using several retrieval strategies. Our results 
show a predictably large drop in performance when models are tested on new knowledge.
Interestingly, when retrieval is introduced, the performance gap between smaller and larger models narrows significantly on knowledge not seen during training.

\section{Related Work}
\label{sec:related-work}

\paragraph{Evaluating Retrieval-Augmented Generation (RAG)}
Retrieval-augmented generation (RAG) is a key strategy to equip large language models (LLMs) with up-to-date information by retrieving relevant external documents at inference time. However, existing RAG evaluations are often undermined by \emph{data contamination}, where evaluation examples overlap with the model’s pretraining corpus. This allows models to answer without retrieval, simply relying on memorized content \citep{li-etal-2024-open-source}. Prominent QA datasets such as Natural Questions \citep{kwiatkowski-etal-2019-natural}, TriviaQA \citep{joshi-etal-2017-triviaqa}, and HotpotQA \citep{yang-etal-2018-hotpotqa} are sourced from Wikipedia or the open web, making it likely that models already ``know'' the answers. This undermines robust assessment of retrieval: models can appear strong simply by regurgitating seen content, and including training examples in prompts can further inflate performance \citep{wang-etal-2022-training}. As a result, current benchmarks struggle to test whether models can truly retrieve and reason over novel information. 

\begin{figure*}[h]
    \centering
    \includegraphics[width=\linewidth]{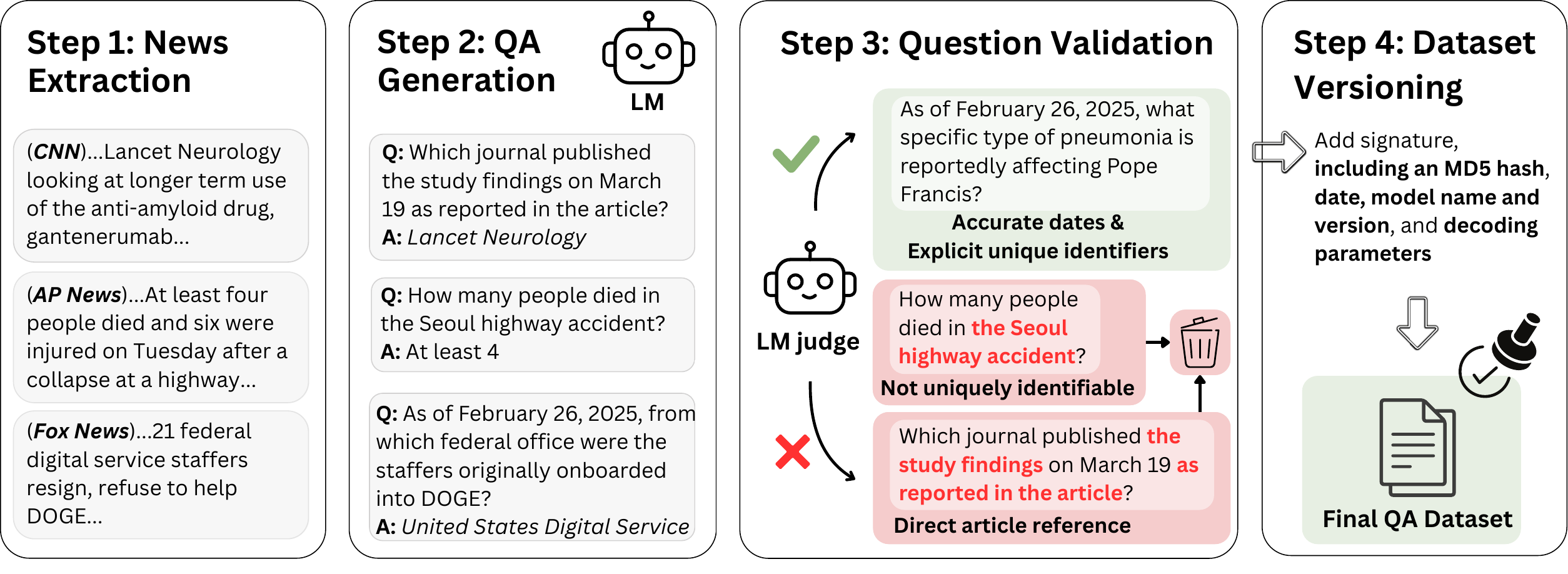}
    \caption{Automated dynamic knowledge benchmark construction pipeline for \name.}
    \label{fig:pipeline}
\end{figure*}

\paragraph{Dynamic Knowledge Question Answering}
To address the limitations of static benchmarks, recent work has introduced \emph{dynamic} QA benchmarks that reflect the evolving state of world knowledge.\footnote{For detailed descriptions of each benchmark, see Appendix~\ref{sec:detailed_benchmarks}.}
StreamingQA \citep{liška2022streamingqabenchmarkadaptationnew} organizes questions chronologically over years of news data, but it does not support continual updates. RealTime QA \citep{kasai2024realtimeqawhatsanswer} delivers weekly quizzes sourced from current news headlines, though coverage is limited by its limited breadth of news feeds and manual curation. FreshQA \citep{vu-etal-2024-freshllms} is a centrally maintained benchmark of roughly 600 author‑ and freelancer‑written, time‑sensitive questions whose answers are periodically updated through extensive human annotation, making its ongoing maintenance costly and dependent on a single coordinating team.

\paragraph{Dynamic Benchmarking Beyond QA}
Dynamic evaluation extends beyond question-answering to other formats.
LiveCodeBench \citep{jain2024livecodebenchholisticcontaminationfree} harvests fresh programming‑contest tasks to build a contamination‑free, time‑stamped suite for code generation, self‑repair, execution, and test prediction; however, public releases so far reach only 2024‑06‑01 and still rely on semi‑manual update scripts. 
LiveBench \citep{livebench} is a challenging, contamination-limited LLM benchmark across multiple reasoning domains and offers partial updates monthly to guarantee a fresh suite of questions bi-yearly.
AntiLeakBench \citep{wu-etal-2025-antileakbench} focuses on preventing data contamination by automatically constructing benchmarks sourced from Wikipedia, but it is constrained to Wikipedia updates.
DeepScholar-Bench \citep{patel2025deepscholarbenchlivebenchmarkautomated} focuses on automatic evaluation for generative research synthesis.
Besides factual evaluation on recent knowledge, automatic benchmarks like Daily Oracle \citep{dai2024llmsprescientcontinuousevaluation}, ForecastBench \citep{karger2025forecastbench} and FutureX \citep{zeng2025futurexadvancedlivebenchmark} test models on the task of forecasting near-future events.
However, forecasting tasks aren't suitable for evaluating retrieval-based methods, as there's often no ground truth database to retrieve from. 
Therefore our focus is factual knowledge from the recent past.

\Cref{tab:dynamic-benchmarks} compares \name{} to other benchmarks along several dimensions. 
In summary, existing dynamic knowledge benchmarks still involve at least partial human curation, infrequent updates, or a narrow focus, and none offer a fully automated, large-scale solution for real-time factual knowledge evaluation.  To our knowledge, \name{} is the first fully automated benchmark for evaluation of factual question answering ability.

\section{Automated Dynamic Benchmarking with \name} 

\subsection{Benchmark Construction Pipeline}
\label{sec:benchmark-pipeline}

We design an agentic framework for dynamic knowledge benchmarking. The pipeline consists of four steps: (1) News extraction, (2) QA generation, (3) question validation, and (4) dataset versioning. An overview of the pipeline is shown in Figure~\ref{fig:pipeline}.

\textbf{News Extraction} \ \
We collect and preprocess news articles published within the past 24 hours from a diverse set of outlets, including both mainstream and specialized publications. The categorization and considered sources of news are presented in Table~\ref{tab:news-sources}. Articles are retrieved via RSS feeds and parsed. For each article, we retain a structured representation that includes metadata such as the title, publication date, author, content body, and source URL.
The output of this step is a curated, timestamped feed of news articles, which serves as the raw knowledge base for dynamic benchmark construction in subsequent stages.

\textbf{QA Generation}  \ \
We use an LLM‑based agent to generate initial multiple‑choice question–answer pairs from curated news articles. The final questions can be delivered in either multiple‑choice or open‑ended format. The agent is instantiated using an LLM\footnote{We use \texttt{gpt-4.1-2025-04-14} in our pipeline.} guided by a prompt designed to elicit high-quality, time-sensitive questions (see \Cref{app:mcq-prompt}). The generation process involves identifying salient facts from each article, drafting a corresponding question, and producing one correct answer along with plausible distractor options. The agent is instructed to prioritize recent and unique facts, particularly entities, events, and developments that are unlikely to appear in older training data. 

\textbf{Question Validation} \ \ 
Despite detailed prompting, LLM-generated questions may not always be well suited for reliable model evaluation. 
In particular, the question sometimes explicitly refers to ``the article,” which is undesirable because we want every question to stand alone.

\begin{table}[t]
\caption{News sources used for dynamic knowledge extraction.}
\label{tab:news-sources}
\centering
\resizebox{1.0\textwidth}{!}{%
\begin{tabular}{@{}ll@{}}
\\
\toprule
\textbf{Category} & \textbf{Sources} \\ \midrule
General / Mainstream News & CNN, BBC, Reuters, The Guardian, Fox News, NBC News, USA Today, HuffPost, CBS News \\
International Coverage & Al Jazeera, DW, RT, Channel News Asia (CNA), Times of India, South China Morning Post (SCMP) \\
Political Focus & Politico, The Hill, NPR \\
Technology and Science & TechCrunch, The Verge, Engadget, Ars Technica, Gizmodo, PC Gamer, TechRadar \\
Business / Finance & Bloomberg \\
Lifestyle / Culture & GQ, Vanity Fair \\
Open-Source Community News & WikiNews \\ \bottomrule
\end{tabular}
}
\end{table}

To address this, we introduce a dedicated question validation agent (see validation prompt in \Cref{app:mcq-prompt}) that assesses the quality and clarity of each question. The agent is tasked with verifying whether each question can be answered uniquely and unambiguously. 

Specifically, it checks whether the question: (1) avoids direct references to the source article, (2) includes accurate and clear date references, (3) uses explicit identifiers for entities such as people, organizations, or events, and (4) avoids vague or ambiguous phrasing. Questions that fail any of these criteria are automatically discarded. Some example QA pairs created by our pipeline are shown in \Cref{tab:example_qa_pairs}.

\textbf{Dataset Versioning} \ \
To support reproducibility and fair comparison, each benchmark release is assigned a unique \emph{signature} serving as its version identifier. Because dataset content can shift due to changes in daily news and the inherent stochasticity of LLM generation, we adopt a versioning approach inspired by SacreBLEU’s reproducibility framework \citep{post-2018-call}. Each signature encodes the agent LLM model name and version (e.g., “gpt-4.1-2025-04-14”), the decoding hyperparameters (temperature, top-$p$, etc.), the dataset generation date and timestamp, and a randomly generated hash (specifically, MD5) as a unique identifier.

Users reporting results on our benchmarks should explicitly cite the full dataset signature and share the corresponding dataset snapshot. This enables precise reproduction and fair evaluation by others.
By versioning each dataset and requiring explicit references, future work can reliably evaluate on the same benchmark instance, which is an essential safeguard in our decentralized benchmarking protocol, since numerous independently generated datasets may potentially exist.

\begin{table*}[ht]
\caption{Example generated QA pairs. The date of dataset generation is February 26, 2025.}
\label{tab:example_qa_pairs}
\centering
\resizebox{0.95\textwidth}{!}{%
  \begin{tabular}{p{6.5cm} p{4.5cm} p{3cm}}
    \\
    \toprule
    \textbf{Question} & \textbf{Choices} & \textbf{Ground Truth} \\
    \midrule
    As of February 26, 2025, what percentage of GDP has UK Prime Minister Keir Starmer announced the country will spend on defense?
    & 
    A. 2.3\% of its GDP 

    B. 3\% of its GDP 

    C. 2.5\% of its GDP 

    D. 7\% of its GDP
    & 
    C. 2.5\% of its GDP\\
    \midrule
    On February 14, 2025, at which hospital was Pope Francis hospitalized for a respiratory infection?
    & 
    A. St.\ Peter's Hospital 

    B. Vatican Medical Center

    C. Gemelli Hospital 

    D. Apostolic Palace Clinic
    & 
    C. Gemelli Hospital  \\
    \midrule
    In which year did Pope Francis have a piece of one lung removed?
    & 
    A. 1967 

    B. 1955 

    C. 1947

    D. 1957
    & 
    D. 1957 \\
    \midrule
    On February 26, 2025, which individual from the Department of Psychiatry at the University of Cambridge emphasized the urgent need for new dementia treatments?
    & 
    A. Dr.\ Marc Siegel 

    B. Dr.\ Ben Underwood 

    C. Dr.\ Chris Vercammen 

    D. Melissa Rudy
    & 
    B. Dr.\ Ben Underwood \\
    \midrule
    As of March 22, 2025, which journal published the study findings on March 19 that detailed the impact of gantenerumab on delaying Alzheimer’s symptoms?
    & 
    A. The Lancet Psychiatry 

    B. JAMA Neurology 

    C. Neurology

    D. The Lancet Neurology
    & 
    D. The Lancet Neurology \\
    \bottomrule
  \end{tabular}%
}
\end{table*}

\begin{table*}[h]
\caption{Example questions that do not pass the human evaluation.}
\label{tab:example_qa_pairs_wrong}
\centering
\resizebox{0.9\textwidth}{!}{%
  \begin{tabular}{p{6.5cm} p{4cm}  p{3cm}}
    \\
    \toprule
    \textbf{Question} & \textbf{Choices} & \textbf{Rationale} \\
    \midrule
    On what date was Pope Francis admitted to the hospital as of March 22, 2025?
    & 
    A. Feb. 7 

    B. Feb. 14  

    C. Feb. 15 

    D. March 22
    & 
    Ambiguity; might rely on past information \\
    \midrule
    When was George Foreman born?
    & 
    A. September 24, 2011

    B. October 30, 1974
    
    C. January 10, 1949

    D. June 3, 2009
    & 
    Rely on past information  \\
    \midrule
    Which streaming service is associated with Severance and previously known for hosting Ted Lasso, as of March 21, 2025?
    & 
    A. Amazon Prime Video

    B. Apple TV+

    C. Hulu

    D. Netflix
    & 
    Rely on past information \\
    \midrule
    As of March 22, 2025, updated HPV shots protect against how many strains of the virus?
    & 
    A. nine 

    B. two

    C. seven

    D. eleven
    & 
    Ambiguity and might rely on past information \\
    \bottomrule
  \end{tabular}%
}
\end{table*}

\subsection{Human Validation}
\label{sec:human-eval}

To evaluate the quality of the generated QA pairs, we ran a 
human validation study on a set of pipeline outputs with two evaluation aspects:
\begin{itemize}[leftmargin=*]
    \item \textbf{Question Quality Check:} Does the question meet the clarity and unambiguity criteria?
    \item \textbf{Answer Correctness Check:} Does the provided correct option exactly match the source article?
\end{itemize}
\noindent
Two independent panels of four computer‑science PhD students (all native‑ or near‑native English speakers) carried out the evaluations. 
The first panel is for evaluating question quality, where each evaluator assessed $60$ multiple-choice questions.
To better control for agreement, $20$ questions in each annotator's part were simultaneously evaluated by $2$ other annotators, each person $10$ questions.
Based on question clarity, the average correctness rate is $\mathbf{92\%}$ over $4$ annotators on $200$ questions in total. \Cref{tab:example_qa_pairs_wrong} shows some QA pairs that did not pass human evaluation.
In the second panel, each evaluator independently rated answer correctness out of 25 QA pairs, and achieved \textbf{100\%} correctness in the sampled questions.

The complete annotation guidelines and survey interfaces used in this study are provided in \Cref{sec:human-annotation-guidelines} and \Cref{sec:human-annotation-interface}. The complete evaluation results are in \Cref{sec:human_annotation_results}.
Because we aim for fully automated, decentralized usage, a small level of noise is acceptable to maintain scalability, freshness, and real-time evaluation.  We also release a daily version of all news collected, enabling on-demand dataset generation under evolving knowledge conditions.  As proprietary LLMs change over time, we will do periodic audits and updates to maintain consistent quality.  By keeping human validation separate from the core pipeline, our framework remains cost‐effective and adaptive, while still supporting quality control when needed.

\subsection{Dataset Statistics and Cost Estimation}
\label{sec:dataset-stats}

Our pipeline ingests news articles from the previous 24~hours and typically yields \(\sim\!\!2{,}000\) multiple-choice questions per run.  
For example, the snapshot generated on 22~March~2025 contains 2{,}350 questions.
\begin{figure*}[!ht]
    \centering
    \includegraphics[width=\linewidth]{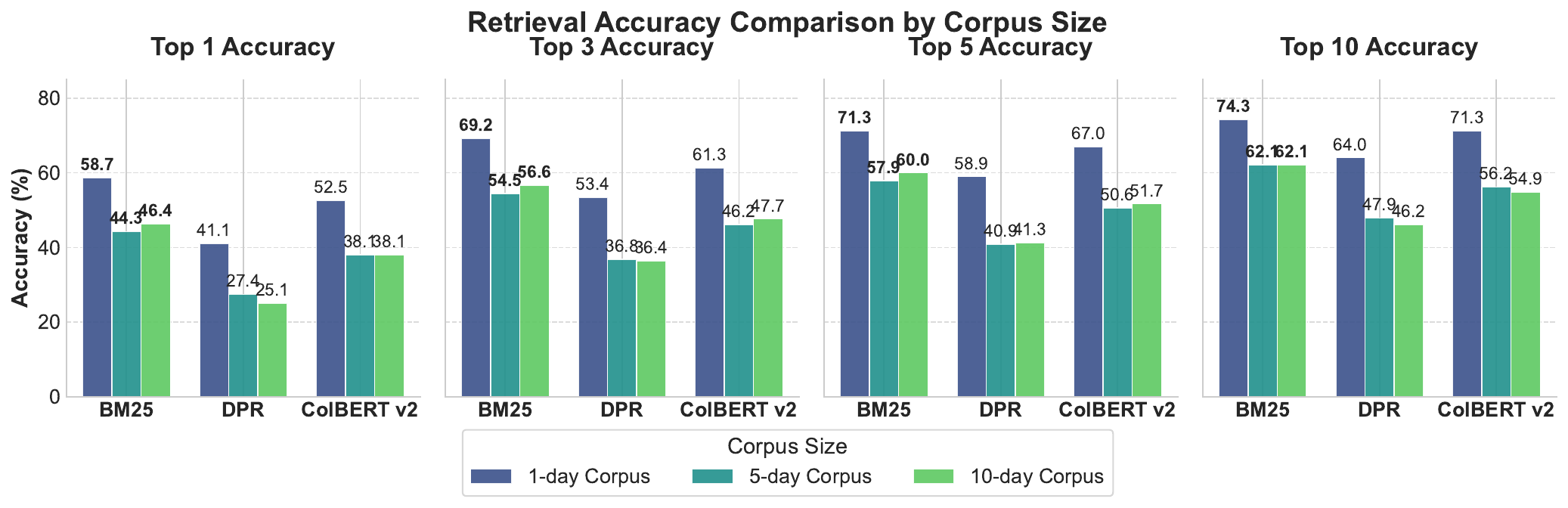}
    \caption{\textbf{Top-$k$ Retrieval Accuracy} for BM25, DPR, and ColBERT v2 across news corpora of different time windows (1-day, 5-day, and 10-day).}
    \label{fig:retrieval_bar}
\end{figure*}
The end‑to‑end expense is modest: generating 2{,}350 raw questions with GPT‑4.1‑2025‑04‑14 costs \(\$2.48\). Validating the same 2{,}350 questions with the model costs an additional \(\$1.73\). Consequently, a full daily benchmark costs roughly \(\$4.21\), making on‑demand generation practical for continuous evaluation.

\subsection{Question Formats}
\label{subsec:question-format}

While the primary format of \name{} is multiple‑choice questions, our pipeline can also generate an open‑ended variant: for each article the generation agent poses a factoid question whose answer is a short span (\(\le\!10\) tokens) copied verbatim from the article; during evaluation we let the model produce up to 100 tokens and pass its first non‑empty line to a separate LLM judge, which simply checks string equality (after normalising case and punctuation) against the ground‑truth span—that span being identical to the correct option in the MC version—and returns a binary correctness decision.

\section{LLM evaluation experiments with \name}
\label{sec:experimental-setup}
In the following experiments, we evaluate a set of models on the \emph{March 22 snapshot} of the dataset
(\Cref{sec:dataset-stats}).\footnote{We focus on this single-day snapshot to provide a concrete, recent evaluation,
though our framework can generate new benchmarks daily. } We evaluate a variety of open-source and proprietary LLMs. For the full list of models, please see \Cref{tab:final-qa-accuracy-simple}.

\paragraph{Evaluation Settings}
We test each LLM under three information-access paradigms:
\begin{enumerate}[label=(\roman*), leftmargin=20pt, itemsep=0pt, topsep=2pt]
    \item \textbf{No-Context}: The model sees only the question and answer choices. We simply provide the prompt: \emph{``Question: \{\texttt{Q}\}. Provide the most accurate answer.''} This reflects a purely parametric recall scenario, where the model must rely solely on its memorized knowledge.

    \item \textbf{Oracle-Context}: The model is given the ground-truth article (i.e., the document originally used to generate the question) as additional context. Here, the model input is of the form: \emph{``Context: \{\texttt{Article}\}. Question: \{\texttt{Q}\}.''} 

    \item \textbf{Retrieval}: We simulate a scenario where the model queries a recent news corpus and retrieves relevant articles before answering. We provide the top-$k$ passages (where $k \in \{1,3,5,10\}$) returned by a retrieval system, concatenated into the prompt. The corpus is drawn from the last 24 hours (1-Day), the preceding 5 days (5-Day), or the preceding 10 days (10-Day). As the corpus grows, more outdated or irrelevant content is introduced, increasing retrieval difficulty.

\end{enumerate}

\paragraph{Retrieval Methods}
We implement a variety of retrievers to provide context in the Retrieval setting. Each daily snapshot of news is indexed using \textbf{BM25 (lexical)} \citep{10.1561/1500000019}, a classic inverted-index-based method leveraging term frequency and inverse document frequency; \textbf{ColBERT v2 (dense)} \citep{santhanam2022colbertv2effectiveefficientretrieval}, which encodes queries and documents at the token level and only then matches each query token to its most similar document token;
and \textbf{DPR (dense)} \citep{karpukhin2020densepassageretrievalopendomain}, a dual-encoder approach producing a single embedding per document and question, scored via dot product. For both dense retrievers, we use FAISS \citep{douze2025faisslibrary} with a flat index for approximate nearest neighbor search. We measure top-$1$, top-$3$, top-$5$, and top-$10$ retrieval accuracy (the fraction of queries where the ground-truth article is among the top-$k$ retrieved documents), as well as final QA performance after the model consumes those retrieved contents.
\begin{figure*}[!t]
    \centering
    \includegraphics[width=\linewidth]{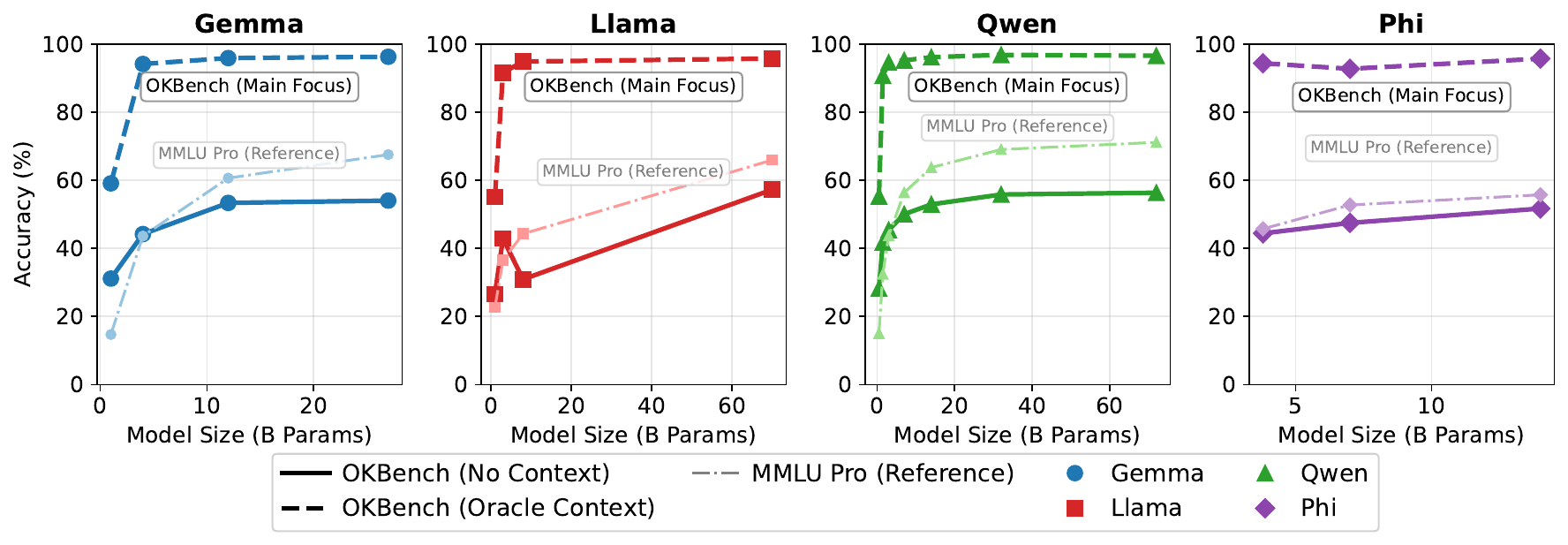}
    \caption{\textbf{No-Context vs.\ Oracle-Context QA Accuracy on \name}, plotted alongside each model's performance on MMLU Pro (lighter lines) as a reference for memorized knowledge.
    We show three representative model families (Gemma, Llama, Qwen) at various parameter scales (Billion Parameters). 
    Solid lines denote \emph{No-Context} accuracy (fresh knowledge), and dashed lines denote \emph{Oracle-Context} accuracy when the ground-truth article is provided. }
    \label{fig:three_panel_accuracy}
\end{figure*}

\section{Evaluation Results and Discussion}
\label{sec:experiments}
\subsection{LLM Knowledge vs.\ Oracle-Context}
Figure~\ref{fig:three_panel_accuracy} summarizes the performance of four representative model families (Gemma, Llama, Qwen, Phi) on \name~in both No-Context and Oracle-Context settings. 
Table~\ref{tab:final-qa-accuracy-simple} in \Cref{sec:appendix-model-results} provides results for a more complete set of models. In addition to multiple-choice questions, we also report open-ended question-answering results in \Cref{sec:appendix-model-results_open_ended}. We find that the open‑ended variant of the benchmark shows a trend similar to the multiple‑choice variant, so we focus on the multiple‑choice version here.

\paragraph{Observation 1: Impact of Fresh Knowledge.} When models must rely solely on parametric memory (No-Context), their performance is far from perfect across all sizes. This reflects the challenge of truly new facts that arise after the model’s pretraining cutoff. Nevertheless, larger models do retain a slight edge. For instance, \texttt{gemma-3-1b-it} only achieves 31.1\% accuracy in No-Context mode, whereas \texttt{gemma-3-27b-it} reaches 54.0\%. The same trend appears in other families like Llama (26.6\% vs.\ 57.2\%) and Qwen (28.2\% vs.\ 56.3\%) when comparing the smallest and largest variants. 
In No-Context mode, Some “fresh‑knowledge’’ questions still reference ongoing stories, such as an election that has been in the news for months, so even a small model can draw on background it has already seen and score above the 25\% random‑guess level. Bigger models possess an even richer store of that prior context, which is why they outperform the smaller ones despite the questions targeting very recent facts.

\paragraph{Observation 2: Oracle-Context and a ``Cutoff'' for Reading Comprehension.} Once the ground-truth article is appended to the query, 
most models (above a certain size threshold) quickly climb to high accuracy (\(\sim95\%\)). 
Even a 4--7~B parameter model can answer correctly given the right passage, 
suggesting that \emph{timely, precise} context is the main determinant of success. 
These findings underscore that for fresh or real-time information, building robust retrieval pipelines may be more critical than simply scaling up model size.

However, contrary to the idea that \emph{all} models do well with the article, 
Figure~\ref{fig:three_panel_accuracy} and Table~\ref{tab:final-qa-accuracy-simple} in \Cref{sec:appendix-model-results} show a sharp performance \textit{cutoff}. 
Models around or above roughly 3--4~B parameters can read and understand the article sufficiently to push their Oracle accuracy to 90--95\%. 
Yet \emph{very small} LLMs (e.g., $\leq$ 1~B parameters) achieve only around 55--60\% even with the ground-truth article. 
This indicates a bound on reading comprehension capacity for extremely small models: 
they simply lack the representational power to parse the passage and correctly pinpoint the answer.

\paragraph{Observation 3: Model Size Scaling behavior on Fresh Data vs.\ Memorized Knowledge.}
The gap between smaller and larger models in the \emph{No context} setting is smaller than one might expect from standard benchmarks that rely heavily on memorized knowledge. To illustrate this point, we also measured each model’s performance on \textbf{MMLU Pro}, a knowledge-intensive benchmark widely used for assessing factual recall from pretraining. \Cref{fig:three_panel_accuracy} and~\Cref{tab:mmlu-pro-results} (in \Cref{sec:mmlu-pro-appendix}) show that on MMLU Pro, scaling from a 1B to a 27B (or 70B) model often yields improvements of 40--50\% or more; in contrast, for our newly generated QA data, the improvement over the same size range is 20--25\%. 
Therefore, while model scale is critical for memorizing facts during pretraining, its benefits are more limited for \emph{emergent} knowledge.

\begin{figure*}[ht]
    \centering
    \includegraphics[width=0.9\linewidth]{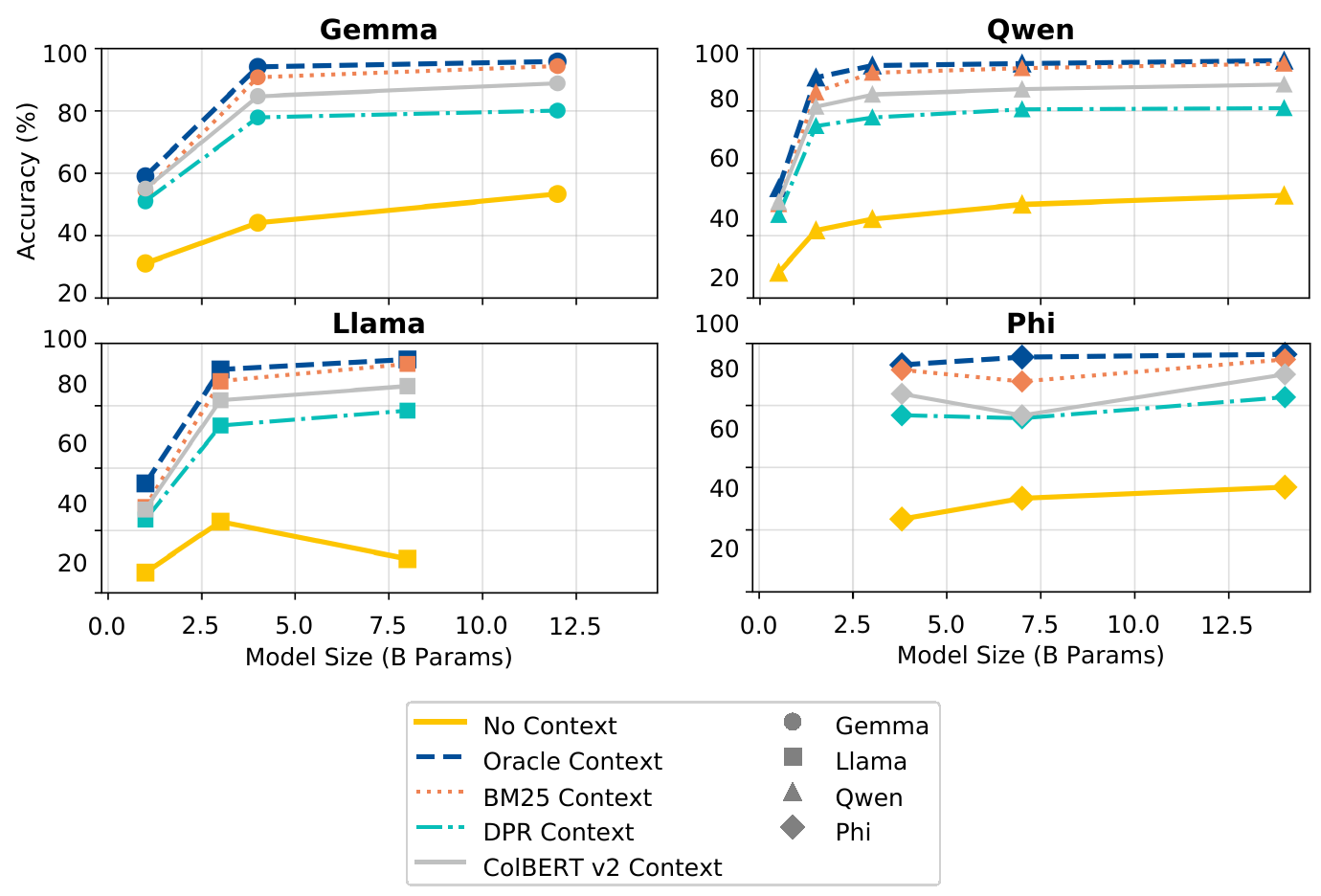}
    \caption{
        \textbf{QA Accuracy with Retrieval-Augmented Context.}
        Each panel shows QA accuracy (\%) for four model families (Gemma, Llama, Qwen, Phi) across different parameter sizes, evaluated on dynamically generated news questions. Lines indicate performance with context retrieved by BM25 (dashed red), ColBERT v2 (solid gray), DPR (dash-dotted cyan), the Oracle (dashed blue, upper bound), and No-Context (solid yellow, lower bound) using top-3 retrieved passages from the 1-day news corpus.
    }
    \label{fig:dynamic_qa_with_retrieval}
\end{figure*}

\begin{table*}[!t]
\caption{Final QA accuracy (\%) of LLMs under Retrieval settings, using
\texttt{Llama-3.1-8B-Instruct} as the QA backbone. Retrieval is performed over 
1-day, 5-day, and 10-day news corpora, returning top-$k$ passages 
($k \in \{1,3,5,10\}$).}
\label{tab:retrieval-qa-accuracy}
\centering
\resizebox{\textwidth}{!}{%
\begin{tabular}{lcccccccccccc}
\\
\toprule
\multirow{2}{*}{\textbf{Retriever}} & \multicolumn{4}{c}{\textbf{1-Day Corpus}} & \multicolumn{4}{c}{\textbf{5-Day Corpus}} & \multicolumn{4}{c}{\textbf{10-Day Corpus}} \\
\cmidrule(lr){2-5}\cmidrule(lr){6-9}\cmidrule(lr){10-13}
& \textbf{Top-1} & \textbf{Top-3} & \textbf{Top-5} & \textbf{Top-10} 
& \textbf{Top-1} & \textbf{Top-3} & \textbf{Top-5} & \textbf{Top-10} 
& \textbf{Top-1} & \textbf{Top-3} & \textbf{Top-5} & \textbf{Top-10} \\
\midrule
BM25       & 90.47 & 93.49 & 93.40 & 92.60 & 88.43 & 91.79 & 92.89 & 92.04 & 88.30 & 91.15 & 92.26 & 92.09 \\
DPR        & 66.26 & 77.66 & 81.28 & 84.21 & 59.49 & 70.89 & 74.34 & 78.13 & 57.53 & 68.60 & 71.57 & 75.96 \\
ColBERT~v2 & 80.09 & 86.13 & 87.79 & 89.32 & 74.17 & 82.55 & 85.02 & 86.43 & 73.06 & 80.72 & 83.49 & 85.45 \\
\bottomrule
\end{tabular}%
}
\end{table*}

\subsection{Retrieval Performance}
\label{subsec:retrieval-analysis}

We experiment with three retrievers: \textbf{BM25}, \textbf{DPR}, and \textbf{ColBERT v2}.
\textbf{Figure~\ref{fig:retrieval_bar}} shows their top-$k$ accuracy on daily news, 
while more detailed numerical results (e.g., top-$1$, top-$3$, etc.) are presented 
in \textbf{Appendix~\ref{app:retrieval-tables}} 
(Tables~\ref{tab:retrieval-hits-accuracy} and~\ref{tab:retrieval-mrr}). Overall, BM25 achieves the highest top-$k$ accuracy in most settings, outperforming both DPR and ColBERT~v2. 

Interestingly, even though dense retrievers like DPR and ColBERT~v2 often excel on standard benchmarks \citep{bajaj2018msmarcohumangenerated, thakur2021beirheterogenousbenchmarkzeroshot}, BM25 proves more robust for this dynamic news scenario. The strong lexical cues (e.g., named entities, event-specific phrasing) may favor exact term matching. It also suggests that domain shift can hurt dense matching unless the models are further adapted, as they are typically trained on MS MARCO \citep{bajaj2018msmarcohumangenerated} (for ColBERT~v2) or Natural Questions \citep{kwiatkowski-etal-2019-natural}(for DPR) rather than on this news domain.

\subsection{Final QA Accuracy with Retrieved Passages}
\label{subsec:retrieval-qa-accuracy}

Finally, we measure how these retrieval methods
impact final question answering performance. \Cref{fig:dynamic_qa_with_retrieval} shows the results using the 1-day news corpus and appending the top 3 retrieved documents across different models (full results in \Cref{sec:appendix-model-results_2}). In line with the earlier retrieval results (cf.\ Figure~\ref{fig:retrieval_bar}),
BM25-based retrieval also yields the highest end-to-end QA performance. 

We also feed the top-$k$ passages 
from each retriever (BM25, DPR, ColBERT~v2) into a moderate-scale 
\texttt{Llama-3.1-8B-Instruct} model and evaluate its QA accuracy. The complete results, including the final QA accuracy (\%) across three corpus sizes (1-day, 5-day, and 10-day) and various $k$ values, are presented in \Cref{tab:retrieval-qa-accuracy}. 
Overall, these results confirm that \emph{accurate retrieval} is vital for time-sensitive QA, perhaps even more so than having a very large model. Even a 1.5B-parameter Qwen model achieves high QA accuracy (above 90\%).
Thus, for ever‑evolving knowledge, robust retrieval pipelines can often compensate for a model’s limited parametric memory.

\section{Conclusion}

We introduce a fully automated framework for dynamic knowledge benchmarking, enabling timely and decentralized evaluation of LLMs. Our agentic pipeline generates high-quality, news-driven QA datasets, supporting robust analysis of model knowledge and retrieval performance. Through experiments on a range of open-source and proprietary models, we demonstrate performance disparities on newly introduced knowledge and the benefits of retrieval augmentation. This work highlights the importance of evaluating LLMs on evolving, non-memorized knowledge to better understand and improve their real-world capabilities.

\section*{Statement on LLM Usage}
We acknowledge the use of Large Language Models (LLMs) to assist in the preparation of this manuscript. Specifically, LLMs were utilized to improve grammar and clarity, aid in literature discovery, and generate boilerplate code snippets for our experiments and testing scripts. The authors have carefully reviewed and edited all LLM-generated outputs and take full responsibility for the final content and scientific integrity of this work.

\section*{Limitations}

\noindent
While our framework democratizes the creation of \emph{dynamic} knowledge benchmarks, several caveats remain:

\begin{itemize}[leftmargin=*]

\item \textbf{Domain \& Language Bias.}  
We currently target English‐language online news.  This excludes non-English, local, pay-walled, or multimedia sources and limits the benchmark’s cultural and topical coverage.  Extending the pipeline to multilingual or domain-specific corpora (e.g., biomedical literature) will require tailored scraping, prompting, and validation strategies.

\item \textbf{Dependence on Proprietary LLMs.}  
Generation and validation agents rely on proprietary frontier models.  Model drift, API quota changes, or access restrictions may affect future reproducibility despite our version-signature protocol.  Moreover, researchers without paid API access may face a cost barrier.

\item \textbf{Legal and Ethical Considerations.}  
We scrape full-text news articles that remain under copyright.  Our release distributes articles for research under fair-use assumptions, but downstream users bear responsibility for local licensing compliance.  Automated harvesting also risks propagating misinformation if upstream outlets publish retracted or false content.

\end{itemize}

\noindent
Addressing these limitations remains important future work for making dynamic knowledge evaluation truly global, robust, and sustainable.

\section*{Acknowledgment}

We thank the Google Gemma Academic
Program for their partial support of Jiawei Zhou
with computational resources.

\bibliography{iclr2026_conference}
\bibliographystyle{iclr2026_conference}

\appendix

\section{Additional Benchmark Details}
\label{sec:detailed_benchmarks}

\paragraph{StreamingQA.}
Builds a time-indexed dataset from a large news corpus (14 years), enabling retrospective testing of how QA models adapt to new information at specific points in history. Once published, it is no longer updated.

\paragraph{RealTime QA.}
Scrapes around 30 weekly questions from news quizzes (e.g., CNN, The Week). Offers a rolling evaluation but is constrained by external quiz sources and weekly time slots, rather than daily updates.

\paragraph{FreshQA.}
Uses a fixed set of around 600 human-written questions whose answers evolve (often involving false premises or rapidly changing facts). Relies on regular human intervention for quality control and updating answers.

\paragraph{Daily Oracle.}
Automatically generates daily forecasting questions (T/F or multiple-choice) from current news, evaluating models' abilities to predict near-future outcomes. Fully automated, but does not focus on post-event factual retrieval or user-driven updates.

\clearpage

\section{Prompt for Generating MCQs}
\label{app:mcq-prompt}

\begin{lstlisting}[basicstyle=\ttfamily\small,breaklines=true]
# News article

**ARTICLE TITLE**:
{article_title}

**ARTICLE TEXT**:
{article_text}

**ARTICLE RELEASE DATE**:
{article_release_date}

# Your task

Generate 5 exceptionally challenging multiple-choice questions based on the article. Follow these requirements:

1. **Question Style**  
   - Use a simple, direct tone. For example:  
     - "Who was elected president of France in 2022?"
     - "Which country hosted the 2023 Climate Summit?"

2. **Question Content**  
   - Each question must focus on factual information about the events or details within the article.  
   - Formulate every question so it can be answered exclusively from the provided content.  
   - Avoid referencing the article directly (do not use phrases like "According to the article..." or "The text indicates...").  
   - For time-sensitive information, incorporate the article's release date. Use "as of {article_release_date}" when referring to ongoing or current information, or "on {article_release_date}" when indicating that an event occurred on that specific day.  
   - Use explicit identifiers for individuals and organizations (e.g., "InfoWars reporter Jamie White"), never ambiguous references like "the official" or "his statement".  
   - Ensure the question is only answerable if one has access to the article (low no-context accuracy).  

3. **Answer Choices**  
   - Provide four (4) plausible choices, each of which is the same entity type (person, organization, place, date, number, etc.).  
   - The correct answer must be an entity present or derivable from the article.  
   - Include distractors that are contextually plausible (either mentioned in the article or logically related).  
   - At least one distractor should closely resemble the correct answer to increase difficulty (e.g., a similar name or date).  
   - Use partial truths or common misconceptions for other distractors, ensuring all choices appear equally plausible without thorough reading.  

4. **Answer Format**  
   - Each question must have a single correct answer (entity) that is taken verbatim from the article.  
   - The answer must not be open-ended: it should be a specific entity (person, organization, place, time, date, number, etc.).  

5. **Question Diversity**  
   - Cover different significant elements or events in the article (avoid repeating the same fact).  
   - Use a variety of question types (who, what, when, where, why, how) and difficulty levels, from moderate to very challenging.  
   - Aim to require different levels of reasoning (recall, inference, analysis).  

6. **Article Release Date**  [IMPORTANT]
   - The article includes a release date provided as `{article_release_date}`. Ensure that this date is incorporated appropriately in questions, using "as of {article_release_date}" for current or ongoing contexts and "on {article_release_date}" when referencing a specific event or fact that happened that day.

7. **Response Format**  
   - Return your final output as a JSON array of exactly 5 objects.  
   - Each object must contain the following keys:
     - `"question_idx"`: An integer from 1 to 5.
     - `"question"`: A string containing the question text.
     - `"choices"`: An array of 4 strings, each a distinct answer option.
     - `"ground_truth"`: A string identical to the correct answer choice from `"choices"`.
     - `"rationale"`: A string explaining why the correct choice is correct and why the others are incorrect.  


Now generate the JSON array with the specified structure:
\end{lstlisting}

\clearpage

\section{Prompt for MCQ Quality Check}
\label{app:mcq-quality-check}

\begin{lstlisting}[basicstyle=\ttfamily\small,breaklines=true]
You are given a multiple-choice question in this format:

{qa_pair}

Check if it meets **all** of the following requirements:

1. **No direct reference to the article**  
   - The question does not begin or contain phrases like "in the article", "According to the article..." or "As reported in the article...".  

2. **Date references are accurate and clear**  
   - If the question references an event or information that took place on a specific date, it can mention that date directly (e.g., "on February 25, 2025").  
   - If the question references a continuing/ongoing situation relative to the article's publication, it should use "as of {article_release_date}" or "on {article_release_date}". 
   - The question should not give ambiguous timing (e.g., "recently" without any date).

3. **Explicit identifiers for individuals or organizations**  
   - Any person or group mentioned must be named clearly (e.g., "The Transportation Ministry" instead of "They" or "That ministry").  
   - Avoid vague references like "the company" or "the government" if a specific entity is known.

4. **No ambiguous references**  
   - If referencing a particular event, location, or study, the question must include all critical details known (e.g., event date, location, or official event name) so that it's clear which event or study is being discussed.  
   - General phrases like "the collapse," "the incident," or "the study" are not acceptable. They must include identifying details such as the location, date, or name.

**Output "1" if *all* the requirements above are met, and "0" otherwise.**
\end{lstlisting}

\clearpage
\section{Human Annotation Guidelines}
\label{sec:human-annotation-guidelines}
\subsection{Question Quality Check}

We are evaluating the quality of a fresh knowledge benchmark dataset designed to test the latest information extracted from up-to-date news articles. This dataset consists of questions, multiple-choice answer candidates, and ground truth answers. Your task is to review the quality of this benchmarking data, specifically checking for clarity and freshness of questions, and the reasonableness of multiple-choice answers based on the provided news.

Please examine both the question and its multiple-choice options for major quality issues. Note that questions were generated on July 16, 2025, based on news articles published on that day.

Here are some guidelines on potential quality issues:

\textbf{Ambiguity}

The question itself should be clear and stand-alone. It should also have an unambiguous answer that is precisely one of the four choices. If you were given a set of articles containing the day's relevant news, you should be able to choose the correct answer. Consider the following:
\begin{itemize}
    \item Questions should be answerable on their own. For example, phrasing like ``\ldots according to the article\ldots'' makes a question unclear.
    \item The time scope of the information requested should be clear. For current information, use phrasing such as ``as of [Date]''; for past events on a given date, use ``on [Date].''
    \item References to events, people, and entities should be clear. For example, use specific names (e.g., ``Dr.\ Ben Underwood''), not vague references like ``the official.'' However, titles or abbreviations are acceptable if they are unambiguous in context (e.g., ``On July 16, 2025, the US President\ldots'').
\end{itemize}

\textbf{Freshness}

We would like questions to be answerable using only the corresponding day's news, not previously known information. Examples of issues that diminish freshness include:
\begin{itemize}
    \item Questions that rely on common sense or widely known facts, and can likely be answered without reading the source article.
    \item Historical trivia, such as questions with static answers (e.g., ``What is Albert Einstein's birthday?'' or ``What was the US population in 2024?''), as these could be answered well before the question date of July 16, 2025.
\end{itemize}

\textbf{Question Quality Decision}

Question: [PLACE HOLDER]\\
Answer Candidate: [PLACE HOLDER]

Please indicate whether the question meets our quality criteria. If you are unsure, make your best guess and provide a comment explaining the uncertainty.
\begin{itemize}
    \item Question passes the quality check
    \item Question is ambiguous/unanswerable
    \item Question is not fresh
    \item Question quality is not good for other reasons (please specify below)
\end{itemize}

\textbf{Any additional comments?}

Please use this space for any comments, such as if your question quality decision is uncertain, or if the question is of poor quality for unlisted reasons.

\subsection{Question Correctness Check}

\textbf{Human Evaluation Instructions}

You will work through 25 multiple-choice questions drawn from 5 different news articles.

The Google Form is organized into 15 sections—three sections for each article—so that you can follow the same three-step procedure every time.

\textbf{Three-step procedure (repeated for every article)}

\begin{enumerate}
    \item Initial guess — prior knowledge only\\
    Read the question without looking at the article or any other source and choose the answer you think is correct.

    \item Article-based answer\\
    Now read the accompanying news article carefully. Based solely on the information in the article, select the option that best answers the question.

    \item Ground-truth check\\
    The ground-truth answer will be shown. Decide whether that answer is exactly supported by the article:\\
    Yes — it matches the article perfectly \textbar{} No — it is contradicted or not stated.
\end{enumerate}

\textbf{Form navigation}

\begin{enumerate}
    \item Complete all three steps for the current article before clicking Next. Once you move to the next section you will not be able to return and edit earlier answers.

    \item Repeat the three-step cycle until you have finished all 15 sections.
\end{enumerate}

Thank you for taking the time to provide careful, accurate responses.

\clearpage
\section{Human Annotation Interface}
\label{sec:human-annotation-interface}

\subsection{Question Quality Check}

\begin{figure*}
    \centering
    \includegraphics[width=\linewidth]{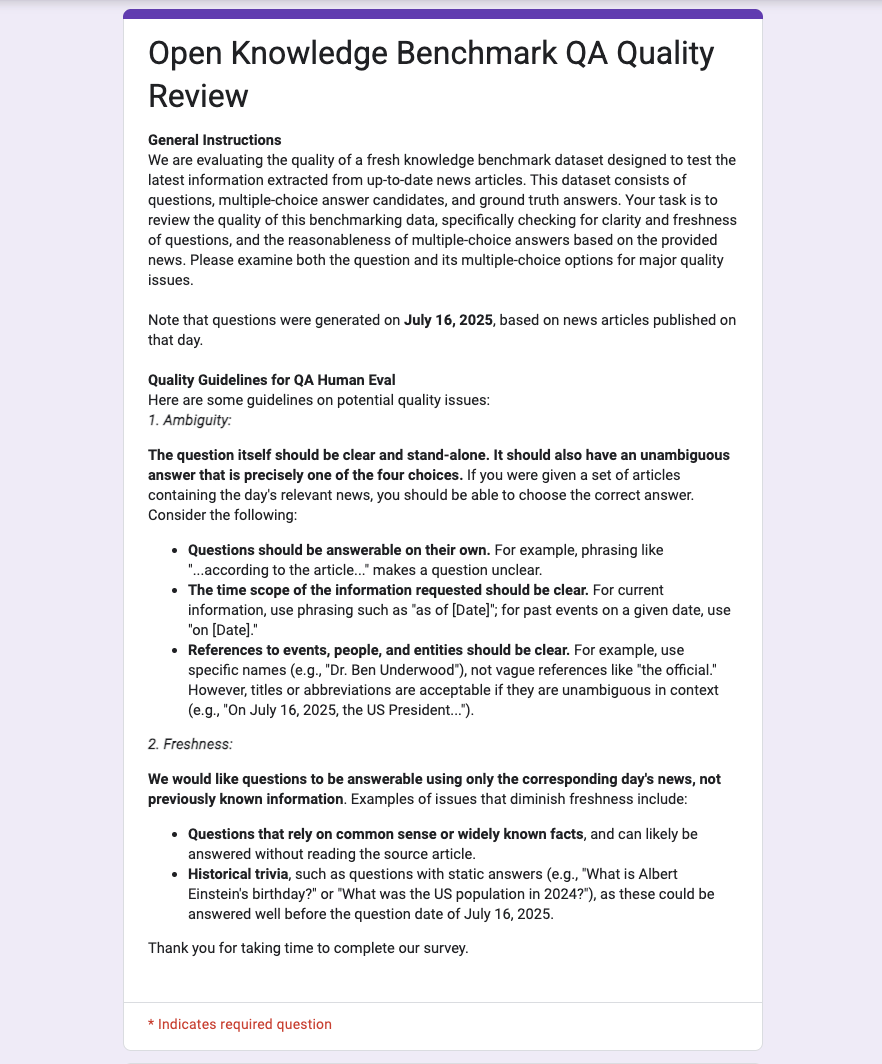}
    \caption{Instructions page of the Google Form used for the Question Quality Check. We ask our human annotators to assess clarity and freshness of the multiple-choice questions, based on the provided instructions.}
    \label{fig:qq_interface_1}
\end{figure*}

\begin{figure*}
    \centering
    \includegraphics[width=\linewidth]{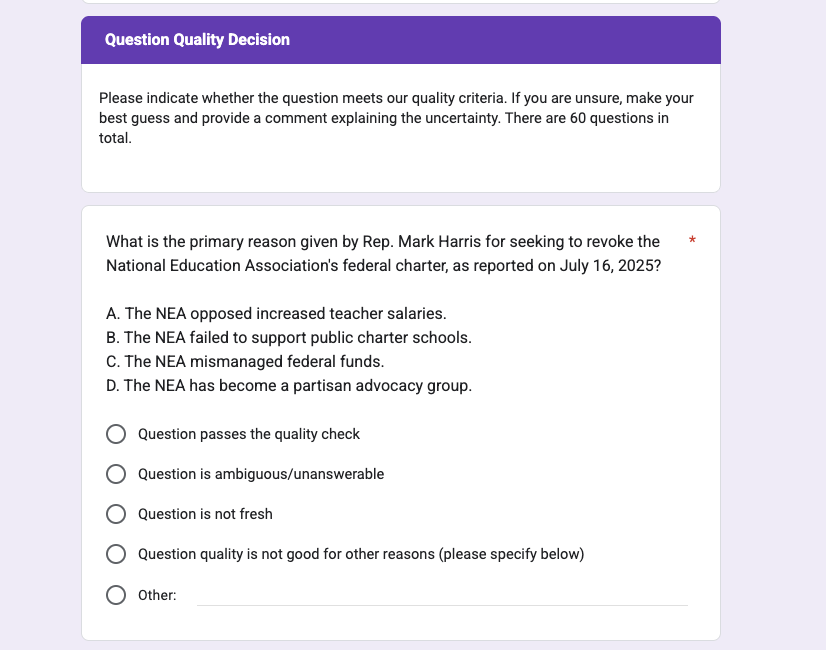}
    \caption{One QA pair example taken from our Question Quality Check survey.}
    \label{fig:qq_interface_2}
\end{figure*}

\begin{figure*}
    \centering
    \includegraphics[width=\linewidth]{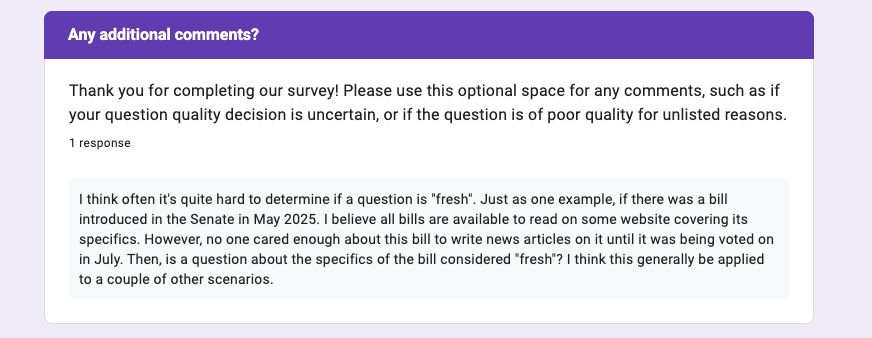}
    \caption{Comments section in our Question Quality Check Survey.}
    \label{fig:qq_interface_3}
\end{figure*}

~\Cref{fig:qq_interface_1,fig:qq_interface_2,fig:qq_interface_3} demonstrates the survey we used to collect human annotation results for quality checking our benchmark. 
We asked $4$ annotators with various backgrounds and interests to each label $60$ questions. 
\Cref{fig:qq_interface_1} is our general instructions which guides our annotators to label the $60$ multiple-choice questions in \Cref{fig:qq_interface_2}. 
We also collects each annotator's feedback after they finish evaluating all $60$ questions to understand human concerns towards our benchmark.

\subsection{Question Correctness Check}

\begin{figure*}
    \centering
    \includegraphics[width=\linewidth]{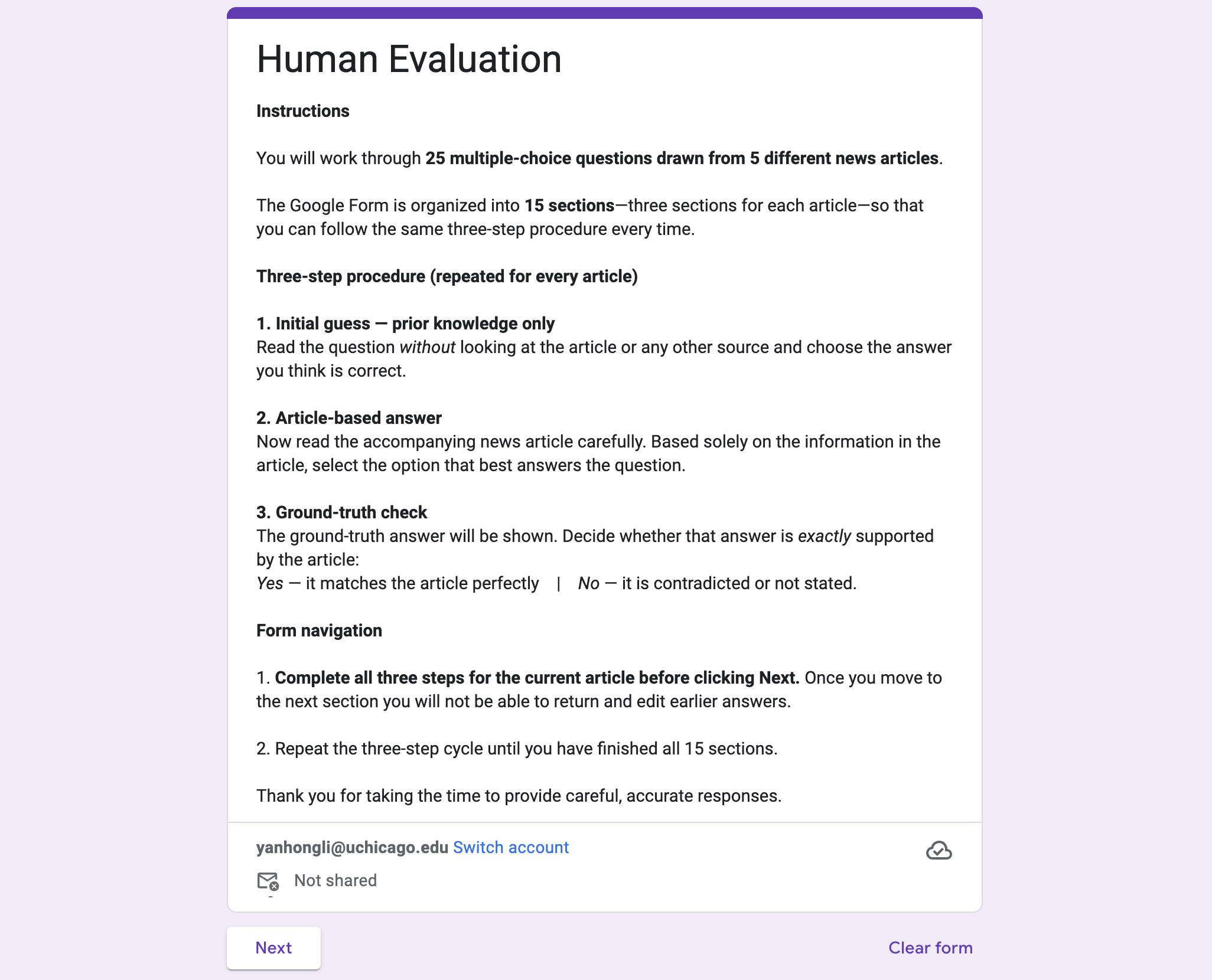}
    \caption{Instructions page of the Google Form used for the Question Correctness Check. It summarises the three‑step workflow and navigation rules for annotators.}
    \label{fig:qc_interface_1}
\end{figure*}

\begin{figure*}
    \centering
    \includegraphics[width=0.8\linewidth]{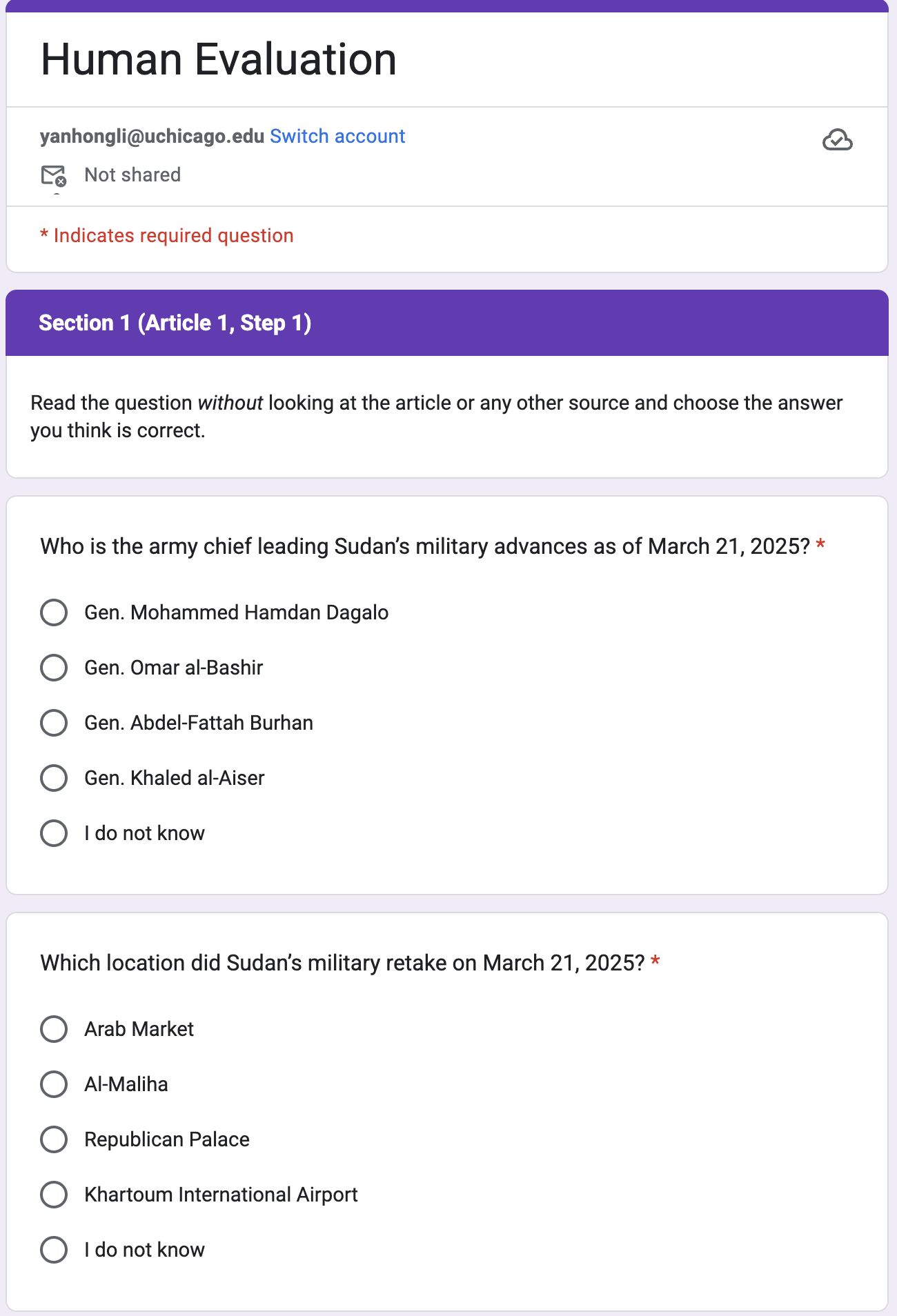}
    \caption{Step 1 (\textbf{Initial guess}). Annotators answer each multiple‑choice question based only on prior knowledge, before seeing the article.}
    \label{fig:qc_interface_2}
\end{figure*}

\begin{figure*}
    \centering
    \includegraphics[width=0.8\linewidth]{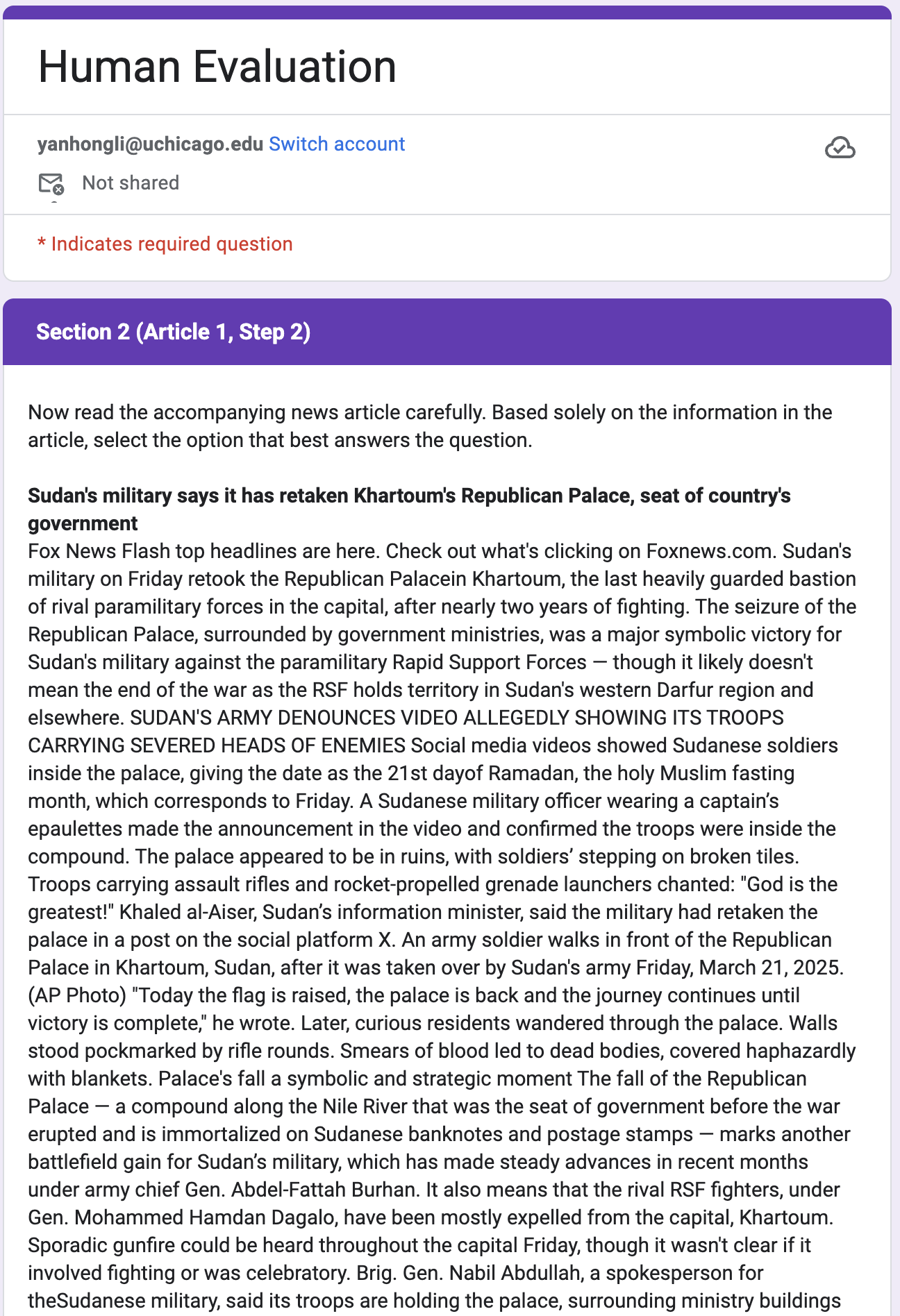}
    \caption{Step 2 (\textbf{Article reading}). The news article is presented in full so annotators can consult it before answering again.}
    \label{fig:qc_interface_3}
\end{figure*}

\begin{figure*}
    \centering
    \includegraphics[width=0.8\linewidth]{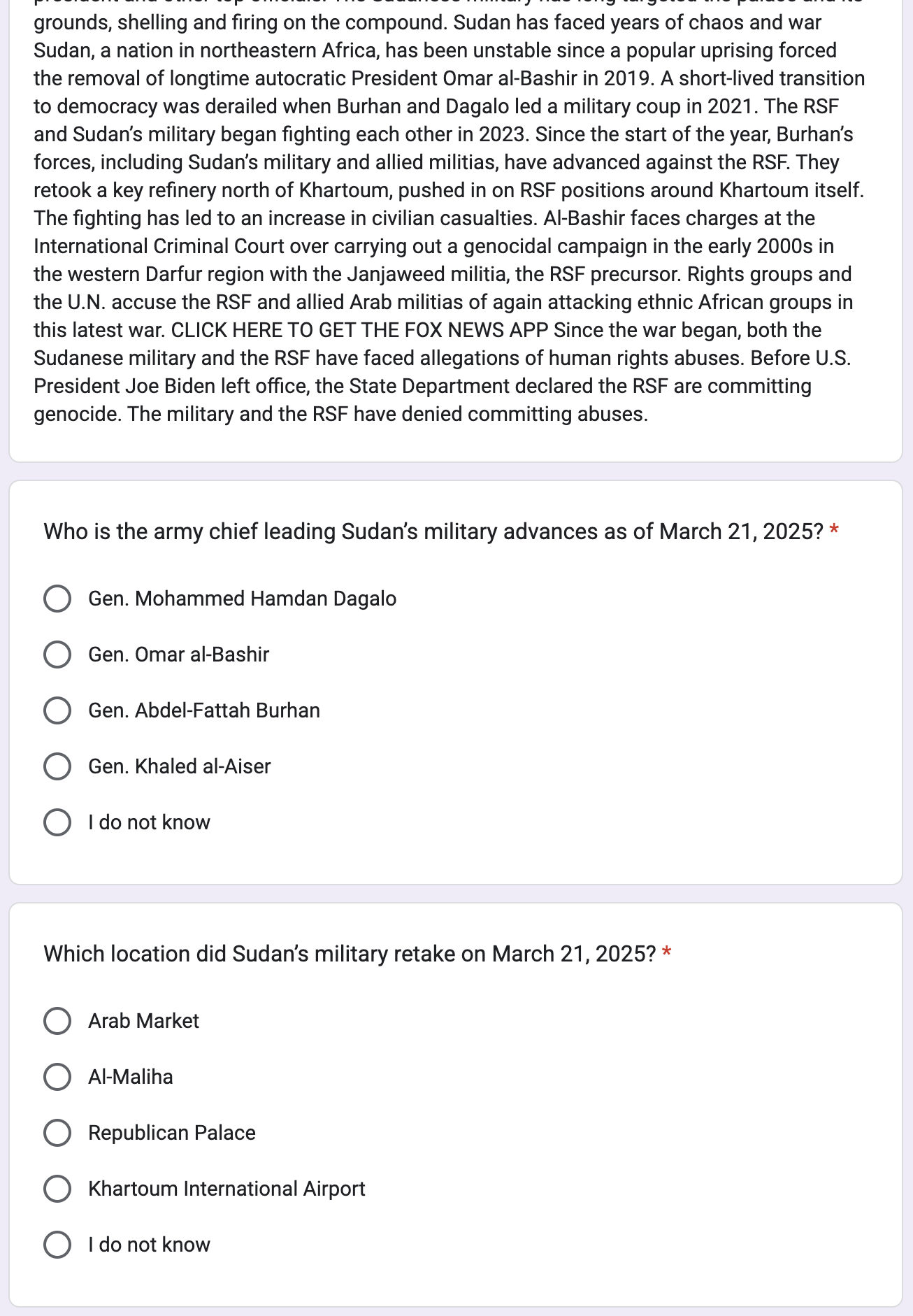}
    \caption{Step 2 (\textbf{Article‑based answer}). After reading the article, annotators choose the option that best answers each question.}
    \label{fig:qc_interface_4}
\end{figure*}

\begin{figure*}
    \centering
    \includegraphics[width=0.8\linewidth]{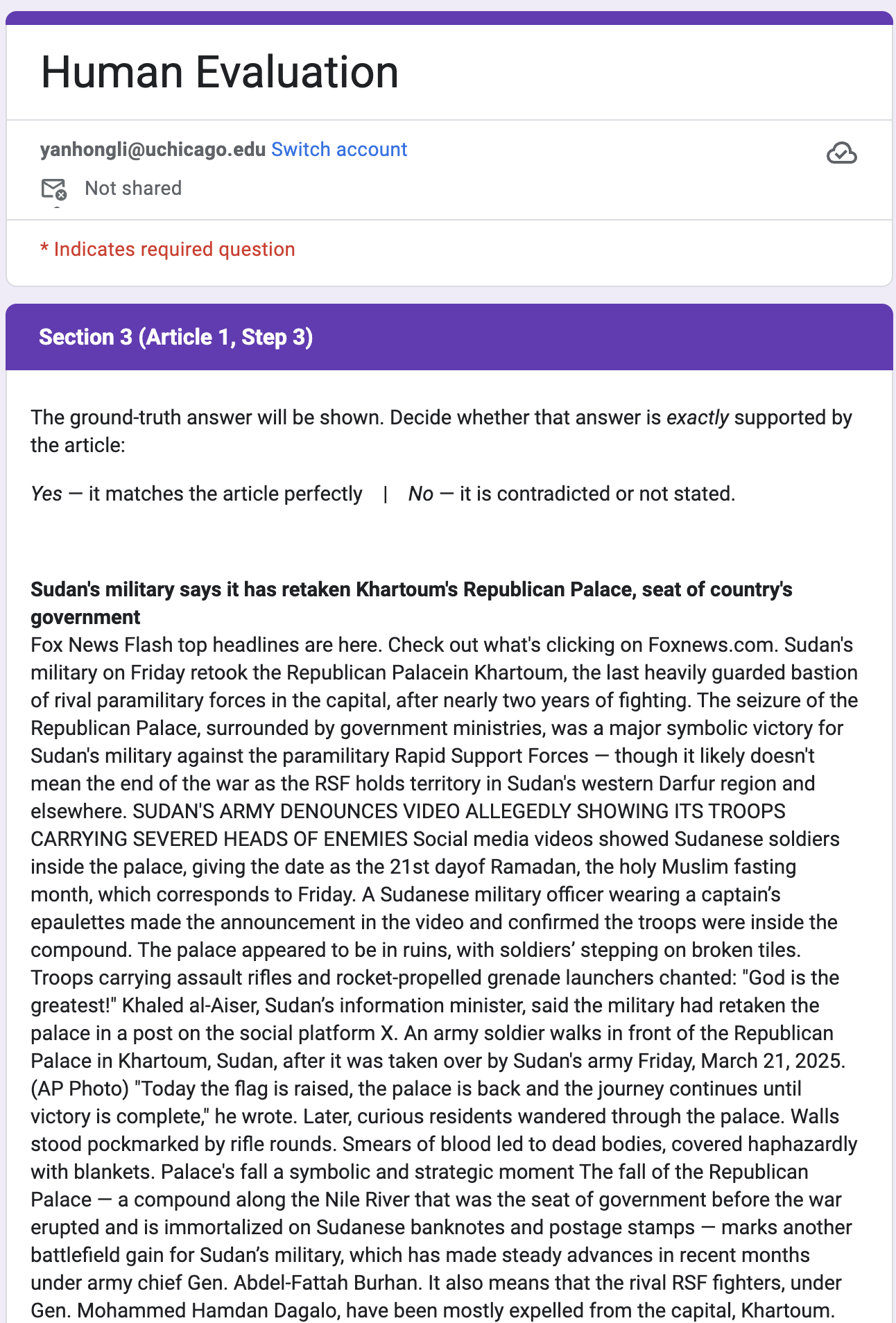}
    \caption{Step 3 (\textbf{Ground‑truth verification instructions}). The form explains how to judge whether the provided ground‑truth answer is exactly supported by the article.}
    \label{fig:qc_interface_5}
\end{figure*}

\begin{figure*}
    \centering
    \includegraphics[width=0.8\linewidth]{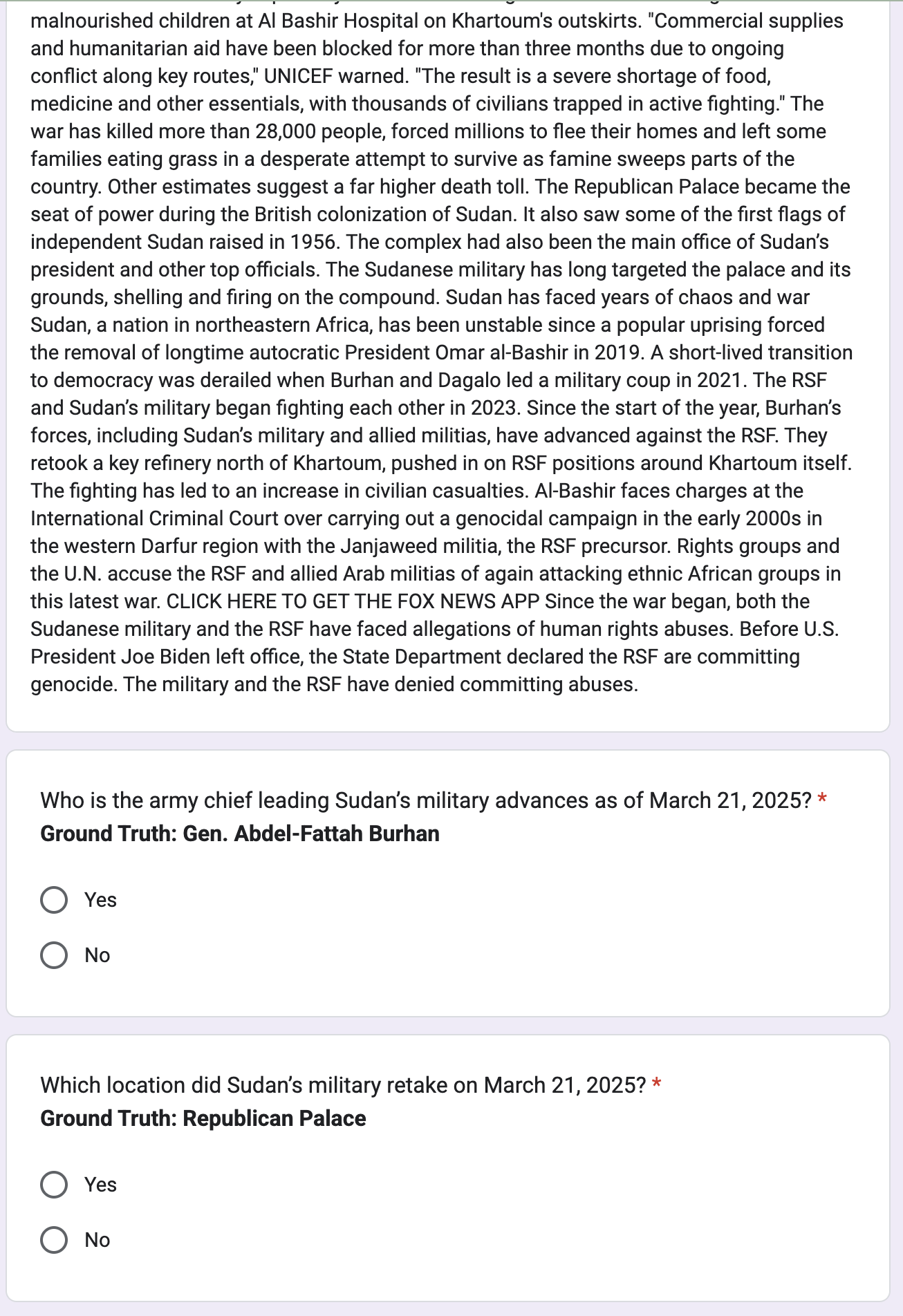}
    \caption{Step 3 (\textbf{Ground‑truth judgment}). Annotators indicate \emph{Yes} if the ground‑truth matches the article or \emph{No} otherwise for each question.}
    \label{fig:qc_interface_6}
\end{figure*}

~\Cref{fig:qc_interface_1,fig:qc_interface_2,fig:qc_interface_3,fig:qc_interface_4,fig:qc_interface_5,fig:qc_interface_6}
walk annotators through the three‑step Google Form we use for the
\emph{Question Correctness Check}.
\Cref{fig:qc_interface_1} presents the instructions page, which explains the task
and navigation rules.
In Step 1, shown in \Cref{fig:qc_interface_2}, annotators make an
\emph{initial guess} for each multiple‑choice question without reading the
article.
Step 2 begins with the full news article (\Cref{fig:qc_interface_3}); after
reading it, annotators answer the same questions again based solely on the
article (\Cref{fig:qc_interface_4}).
Finally, Step 3 displays the ground‑truth answers and asks annotators to judge
whether they are exactly supported by the article
(\Cref{fig:qc_interface_5,fig:qc_interface_6}).

\clearpage
\section{Human Annotation Results}
\label{sec:human_annotation_results}

\subsection{Question Quality Check}
\Cref{tab:qq_results} shows the human evaluation results of \emph{Question Quality Check}.
Across four annotators, the average correctness rate is $92\%$ based on clarity only.
It's noteworthy that our human annotators mostly disagree on freshness, since none of the questions was labeled as ambiguous by both annotators. 
Out of $40$ overlapped questions, there are $2$ questions labeled as not fresh by both annotators, and $1$ question where both annotators labeled as fail out of different reasons.
We observe that assessing question freshness is generally more difficult, due to the limitations of news articles, humans and LLMs.
As noted by one of our annotators, it's challenging to determine if a question is ``fresh'' in multiple scenarios, since some seemingly trivial information from the past may be randomly covered by future news articles.
Therefore, we consider clarity as the primary metric to assess the quality of our benchmark, as freshness can be easily affected by a lot of uncontrollable factors.
When evaluated on both clarity and freshness, the average correctness rate is $66\%$ on $200$ multiple-choice questions.

\subsection{Question Correctness Check}
\Cref{tab:qc_results} summarises the outcomes of the three‑step
\emph{Question Correctness Check}.
Part 1 measures how often annotators could guess the correct answer
\emph{before} reading the article; the low 10 \% accuracy confirms that the
questions are not answerable from prior knowledge alone.
After consulting the article (Part 2), all annotators selected the ground‑truth
multiple‑choice option in every case (100\% agreement), indicating the
questions are clear and the correct choice is recoverable from the article.
In Part 3, annotators judged whether the ground‑truth answer is \emph{exactly}
supported by the article; 88 \% of judgments were “Yes”.
All 12 disputed items were produced by the same annotator, who later acknowledged that they were unsure of the date of the new passage and therefore over‑thought their answers. Therefore, the misunderstanding stems from the questionnaire design, not from the answers being incorrect.

\begin{table*}[t]
    \caption{Human annotation results for the Question Quality Check.
    Each annotator is responsible for labeling $60$ multiple-choice questions, where they assess question clarity and knowledge freshness. 
    For each annotator's part, $20$ out of $60$ questions are annotated by $2$ different annotators to calculate the agreement ratio.
    In total, $200$ questions are evaluated for quality.}
    \label{tab:qq_results}
    \centering
    \resizebox{0.97\textwidth}{!}{%
    \begin{tabular}{lccccc}
        \\
        \toprule
        \textbf{Annotator} &
        \textbf{Question Range} &
        \textbf{Failed Clarity} &
        \textbf{Failed Freshness} &
        \textbf{Correctness} &
        \textbf{Agreement} \\
        &\small indices & \small\# ambiguous questions &
        \small\# outdated questions &
        \small\# passed questions &
        \small\# matched failures \\
        \midrule
        1 & 1-60 & 5 & 4 & 51 & 0/20 \\
        2 & 51-110 & 1 & 11 & 45 & 0/20 \\
        3 & 101-160 & 3 & 26 & 31 & 3/20 \\
        4 & 1-10, 151-200 & 10 & 17 & 31 & 1/20 \\
        \bottomrule
    \end{tabular}
    }
\end{table*}

\begin{table*}[t]
    \caption{Human annotation results for the Question Correctness Check.
    Part 1 assesses prior‑knowledge accuracy,
    Part 2 agreement with the multiple‑choice ground truth after reading the
    article, and Part 3 verification that the ground‑truth answer is fully
    supported by the article.}
    \label{tab:qc_results}
    \centering
    \begin{tabular}{lccc}
        \\
        \toprule
        \textbf{Annotator} &
        \textbf{Part 1} &
        \textbf{Part 2} &
        \textbf{Part 3} \\
        & \small\# correct guesses &
        \small\# matched answers &
        \small\# “Yes” judgments \\
        \midrule
        1 & 1 & 25 & 25 \\
        2 & 1 & 25 & 25 \\
        3 & 2 & 25 & 13 \\
        4 & 6 & 25 & 25 \\
        \midrule
        \textbf{Total (out of 100)} & 10 & 100 & 88 \\
        \bottomrule
    \end{tabular}
\end{table*}

\clearpage
\section{Complete Model Benchmarking Results (Multiple Choice Format)}
\label{sec:appendix-model-results}

Table~\ref{tab:final-qa-accuracy-simple} shows the final QA accuracy (\%) for a broad range of open-sourced and proprietary LLMs under both \emph{No-Context} and \emph{Oracle} settings. 
As discussed in the main paper, these results highlight the importance of timely context for questions involving fresh, real-world information and illustrate a performance ``cutoff'' phenomenon for smaller model sizes (e.g., 1B parameters) versus larger ones (e.g., 7B or more). 
``Oracle'' accuracy steadily approaches near-ceiling for models above roughly 3--4B parameters, indicating a scaling threshold for effective reading comprehension on time-sensitive content.

\begin{table*}[!ht]
\centering
\caption{Final QA accuracy (\%) of open-sourced and closed-sourced LLMs under No-Context and Oracle (Context) settings.}
\label{tab:final-qa-accuracy-simple}
\resizebox{0.7\textwidth}{!}{%
\begin{tabular}{lcc}
\\
\toprule
\textbf{Model} & \textbf{No-Context Acc} & \textbf{Oracle Acc} \\
\midrule
\textbf{Open-Sourced Models} & & \\
\midrule
\texttt{gemma-3-1b-it}             & 31.11 & 59.06 \\
\texttt{gemma-3-4b-it}             & 44.17 & 94.09 \\
\texttt{gemma-3-12b-it}            & 53.32 & 95.83 \\
\texttt{gemma-3-27b-it}            & 54.00 & 96.21 \\
\texttt{Llama-3.2-1B-Instruct}     & 26.55 & 55.06 \\
\texttt{Llama-3.2-3B-Instruct}     & 42.85 & 91.57 \\
\texttt{Llama-3.1-8B-Instruct}     & 30.89 & 94.81 \\
\texttt{Llama-3.3-70B-Instruct}    & 57.23 & 95.70 \\
\texttt{Phi-3-mini-128k-instruct}  & 44.38 & 94.30 \\
\texttt{Phi-3-small-128k-instruct}  & 47.45  & 92.68 \\
\texttt{Phi-3-medium-128k-instruct}  & 51.66  & 95.66 \\
\texttt{Phi-4-mini-instruct}       & 43.57 & 93.62 \\
\texttt{Qwen2.5-0.5B-Instruct}     & 28.17 & 55.19 \\
\texttt{Qwen2.5-1.5B-Instruct}     & 41.70 & 90.64 \\
\texttt{Qwen2.5-3B-Instruct}       & 45.36 & 94.51 \\
\texttt{Qwen2.5-7B-Instruct}       & 50.00 & 95.15 \\
\texttt{Qwen2.5-14B-Instruct}      & 52.89 & 96.09 \\
\texttt{Qwen2.5-32B-Instruct}      & 55.79 & 96.77 \\
\texttt{Qwen2.5-72B-Instruct}      & 56.30 & 96.51 \\
\texttt{Mistral-7B-Instruct-v0.2}       & 35.96 & 90.21 \\
\texttt{Mistral-Small-24B-Instruct-2501} & 53.23 & 96.43 \\
\texttt{Mixtral-8x7B-Instruct-v0.1}      & 33.36 & 93.40 \\
\midrule
\textbf{Proprietary Models} & & \\
\midrule
\texttt{GPT-4o}       & 59.96 & 96.60 \\
\texttt{GPT-o1-mini}  & 32.38 & 96.34 \\
\texttt{GPT-o3-mini}  & 55.36 & 97.28 \\
\texttt{Gemini-1.5-pro} & 55.15 & 96.51 \\
\bottomrule
\end{tabular}%
}
\end{table*}

\clearpage
\subsection{MMLU Pro: Memorized Knowledge Assessment}
\label{sec:mmlu-pro-appendix}

\begin{table*}[ht]
    \centering
    \caption{\textbf{MMLU Pro Results} (\% accuracy). 
    We report performance on a knowledge-intensive QA benchmark, reflecting memorized or static knowledge from pre-training.}
    \label{tab:mmlu-pro-results}
    \begin{tabular}{lcr}
        \\
        \toprule
        \textbf{Model} & \textbf{Size} & \textbf{Accuracy (\%)} \\
        \midrule
        \texttt{Llama-3.2-1B-Instruct}  & 1B   & 22.6 \\
        \texttt{Llama-3.2-3B-Instruct}  & 3B   & 36.5 \\
        \texttt{Llama-3.1-8B-Instruct}  & 8B   & 44.3 \\
        \texttt{Llama-3.3-70B-Instruct} & 70B  & 65.9 \\
        \midrule
        \texttt{Gemma-3-1B}   & 1B   & 14.7 \\
        \texttt{Gemma-3-4B}   & 4B   & 43.6 \\
        \texttt{Gemma-3-12B}  & 12B  & 60.6 \\
        \texttt{Gemma-3-27B}  & 27B  & 67.5 \\
        \midrule
        \texttt{Qwen-2.5-0.5B}  & 0.5B  & 15.0 \\
        \texttt{Qwen-2.5-1.5B}  & 1.5B  & 32.4 \\
        \texttt{Qwen-2.5-3B}    & 3B    & 43.7 \\
        \texttt{Qwen-2.5-7B}    & 7B    & 56.3 \\
        \texttt{Qwen-2.5-14B}   & 14B   & 63.7 \\
        \texttt{Qwen-2.5-32B}   & 32B   & 69.0 \\
        \texttt{Qwen-2.5-72B}   & 72B   & 71.1 \\
        \bottomrule
    \end{tabular}
\end{table*}

In Table~\ref{tab:mmlu-pro-results}, we report the accuracy of various models on the MMLU Pro benchmark, a knowledge-intensive QA dataset aimed at evaluating factual recall from pre-training.
These results offer insight into how well each model retains \emph{static} domain knowledge, in contrast to the \emph{dynamic}, newly emerging facts tested by our daily-updated QA benchmark.
We observe that scaling model size often brings significant improvements in MMLU Pro accuracy, reflecting the growing capacity for memorizing factual content. 
Notably, the performance gains on MMLU Pro can be substantially larger than the gains observed on our fresh-news dataset under No-Context conditions, underscoring the difference between learned “long-term” knowledge and newly introduced facts.

\clearpage
\section{Complete Model Benchmarking Results (Open Ended Questions)}
\label{sec:appendix-model-results_open_ended}

\Cref{tab:final-qa-accuracy-simple-oe} reports the open-ended question‑answering accuracy (\%) of every
model we evaluated under both \emph{No‑Context} and \emph{Oracle} settings.
The table consolidates results for open‑sourced and closed‑sourced LLMs,
making it easy to trace how providing the exact answer‑containing passage
(“Oracle”) closes the gap that appears when models must rely solely on their
parametric knowledge (“No‑Context”).  Reading downward, you can also see the
scale threshold—around 3‑4B parameters—beyond which Oracle accuracy plateaus
near ceiling, while smaller models lag substantially without context.

\begin{table*}[h]
\centering
\caption{Final accuracy (\%) of all tested models in the \emph{No‑Context}
         and \emph{Oracle} settings usin open-ended question format.}
\label{tab:final-qa-accuracy-simple-oe}
\resizebox{0.8\textwidth}{!}{%
\begin{tabular}{lrr}
\\
\toprule
\textbf{Model} & \textbf{No‑Context Acc} & \textbf{Oracle Acc} \\
\midrule
\multicolumn{3}{c}{\textbf{Open‑Sourced Models}} \\
\midrule
\texttt{gemma-3-1b-it}             & 4.64  & 70.09 \\
\texttt{gemma-3-4b-it}             & 10.51 & 86.09 \\
\texttt{gemma-3-12b-it}            & 14.17 & 89.79 \\
\texttt{gemma-3-27b-it}            & 17.19 & 89.79 \\
\texttt{Llama-3.2-1B-Instruct}     & 2.81  & 74.72 \\
\texttt{Llama-3.2-3B-Instruct}     & 5.19  & 86.64 \\
\texttt{Llama-3.1-8B-Instruct}     & 2.68  & 82.94 \\
\texttt{Llama-3.3-70B-Instruct}    & 16.13 & 90.13 \\
\texttt{Phi-3-mini-128k-instruct}  & 8.26  & 82.64 \\
\texttt{Phi-4-mini-instruct}       & 8.98  & 75.19 \\
\texttt{Qwen2.5-0.5B-Instruct}     & 4.47  & 70.94 \\
\texttt{Qwen2.5-1.5B-Instruct}     & 7.02  & 85.53 \\
\texttt{Qwen2.5-3B-Instruct}       & 6.98  & 86.85 \\
\texttt{Qwen2.5-7B-Instruct}       & 9.36  & 89.62 \\
\texttt{Qwen2.5-14B-Instruct}      & 10.43 & 84.68 \\
\texttt{Qwen2.5-32B-Instruct}      & 11.36 & 91.02 \\
\texttt{Qwen2.5-72B-Instruct}      & 13.79 & 90.94 \\
\texttt{Mistral-7B-Instruct-v0.2}       & 5.19  & 84.47 \\
\texttt{Mistral-Small-24B-Instruct-2501} & 12.77 & 90.64 \\
\texttt{Mixtral-8x7B-Instruct-v0.1}      & 8.60  & 86.34 \\
\midrule
\multicolumn{3}{c}{\textbf{Closed‑Sourced Models}} \\
\midrule
\texttt{GPT-4o-2024-08-06} & 17.74 & 91.79 \\
\texttt{o1-mini-2024-09-12} & 4.43 & 88.30 \\
\texttt{o3-mini-2025-01-31} & 15.79 & 92.38 \\
\texttt{Gemini-1.5-pro}     & 17.83 & 90.13 \\
\bottomrule
\end{tabular}}
\end{table*}

\clearpage
\section{Additional Retrieval Results}
\label{app:retrieval-tables}

To provide a fuller picture of how our retriever stack behaves under different temporal scopes, Table \ref{tab:retrieval-hits-accuracy} details the Top‑$k$ hit rates—the fraction of questions whose gold article appears within the first $k$ results—while Table \ref{tab:retrieval-mrr} complements this view with Top‑$k$ mean reciprocal rank (MRR), capturing average ranking quality.  We report both metrics for BM25, DPR, and ColBERT v2 across three corpus sizes (news from the last 1, 5, and 10 days) and four cut‑off values ($k = 1,3,5,10$).  Together, these tables reveal how retrieval effectiveness degrades as the candidate pool widens, and how each method trades off early‑precision (Top‑1/3) versus broader recall (Top‑10) under increasingly challenging settings.

\begin{table*}[!ht]
\centering
\caption{Top-$k$ hits accuracy (\%) for different retrieval methods across 1-day, 5-day, and 10-day corpora. Each cell represents the fraction of questions for which the ground-truth article is ranked within the top $k$ results.}
\label{tab:retrieval-hits-accuracy}
\resizebox{\textwidth}{!}{%
\begin{tabular}{lcccccccccccc}
\\
\toprule
\multirow{2}{*}{\textbf{Retriever}} & 
\multicolumn{4}{c}{\textbf{1-Day Corpus}} & 
\multicolumn{4}{c}{\textbf{5-Day Corpus}} & 
\multicolumn{4}{c}{\textbf{10-Day Corpus}} \\
\cmidrule(lr){2-5}\cmidrule(lr){6-9}\cmidrule(lr){10-13}
& \textbf{Top-1} & \textbf{Top-3} & \textbf{Top-5} & \textbf{Top-10} &
  \textbf{Top-1} & \textbf{Top-3} & \textbf{Top-5} & \textbf{Top-10} &
  \textbf{Top-1} & \textbf{Top-3} & \textbf{Top-5} & \textbf{Top-10} \\
\midrule
BM25       
& 58.72 & 69.15 & 71.28 & 74.26 
& 44.26 & 54.47 & 57.87 & 62.13
& 46.38 & 56.60 & 60.00 & 62.13 \\

DPR        
& 41.06 & 53.40 & 58.94 & 64.04 
& 27.45 & 36.81 & 40.85 & 47.87
& 25.11 & 36.38 & 41.28 & 46.17 \\

ColBERT v2 
& 52.55 & 61.28 & 67.02 & 71.28 
& 38.09 & 46.17 & 50.64 & 56.17
& 38.09 & 47.66 & 51.70 & 54.89 \\
\bottomrule
\end{tabular}%
}
\end{table*}

\begin{table*}[!ht]
\centering
\caption{Top-$k$ Mean Reciprocal Rank (MRR) for different retrieval methods across 1-day, 5-day, and 10-day corpora. Each cell represents the average reciprocal rank of the ground-truth article.}
\label{tab:retrieval-mrr}
\resizebox{\textwidth}{!}{%
\begin{tabular}{lcccccccccccc}
\\
\toprule
\multirow{2}{*}{\textbf{Retriever}} & 
\multicolumn{4}{c}{\textbf{1-Day Corpus}} & 
\multicolumn{4}{c}{\textbf{5-Day Corpus}} & 
\multicolumn{4}{c}{\textbf{10-Day Corpus}} \\
\cmidrule(lr){2-5}\cmidrule(lr){6-9}\cmidrule(lr){10-13}
& \textbf{Top-1} & \textbf{Top-3} & \textbf{Top-5} & \textbf{Top-10} &
  \textbf{Top-1} & \textbf{Top-3} & \textbf{Top-5} & \textbf{Top-10} &
  \textbf{Top-1} & \textbf{Top-3} & \textbf{Top-5} & \textbf{Top-10} \\
\midrule
BM25       
& 0.59 & 0.63 & 0.64 & 0.64
& 0.44 & 0.49 & 0.50 & 0.50
& 0.46 & 0.51 & 0.52 & 0.52 \\

DPR        
& 0.41 & 0.47 & 0.48 & 0.49
& 0.27 & 0.32 & 0.32 & 0.33
& 0.25 & 0.30 & 0.31 & 0.32 \\

ColBERT v2 
& 0.53 & 0.56 & 0.58 & 0.58
& 0.38 & 0.42 & 0.43 & 0.43
& 0.38 & 0.43 & 0.43 & 0.44 \\
\bottomrule
\end{tabular}%
}
\end{table*}

\clearpage

\section{Complete End‑to‑End QA Results}
\label{sec:appendix-model-results_2}

Table~\ref{tab:retrieval-qa-accuracy} reports the final question‑answering
accuracy (\%) when each LM receives the top 3 passages returned by three
retrievers—BM25, DPR, and ColBERT v2—on the 1‑day news corpus.
These numbers complement Figure~\ref{fig:dynamic_qa_with_retrieval} by
revealing how retrieval quality interacts with model size \emph{across the
full model set}.  Higher accuracies for BM25 corroborate our main‑text claim
that lexical cues (named entities, dates) dominate in rapidly evolving news,
while dense retrievers lag unless adapted to the domain.

\begin{table*}[t]
\centering
\small
\setlength{\tabcolsep}{6pt}
\caption{End‑to‑end QA accuracy (\%) on the 1‑day news corpus
with the top‑3 retrieved passages appended to each query.}
\label{tab:retrieval-qa-accuracy_1}
\begin{tabular}{lccc}
\\
\toprule
\textbf{Model} & \textbf{BM25} & \textbf{DPR} & \textbf{ColBERT v2} \\
\midrule
Gemma‑3‑1B‑IT            & 54.43 & 51.06 & 55.06 \\
Gemma‑3‑4B‑IT            & 90.72 & 77.91 & 84.68 \\
Gemma‑3‑12B‑IT           & 94.34 & 80.11 & 88.77 \\
Gemma‑3‑27B‑IT           & 95.28 & 77.19 & 86.77 \\[2pt]
Llama‑3.2‑1B‑Instruct    & 47.49 & 43.53 & 46.77 \\
Llama‑3.2‑3B‑Instruct    & 87.83 & 73.62 & 81.79 \\
Llama‑3.1‑8B‑Instruct    & 93.36 & 78.43 & 86.26 \\
Llama‑3.3‑70B‑Instruct   & 94.98 & 78.13 & 86.98 \\[2pt]
Qwen 2.5‑0.5B‑Instruct   & 50.17 & 46.68 & 50.77 \\
Qwen 2.5‑1.5B‑Instruct   & 85.96 & 75.11 & 81.36 \\
Qwen 2.5‑3B‑Instruct     & 92.17 & 77.87 & 85.23 \\
Qwen 2.5‑7B‑Instruct     & 93.66 & 80.51 & 86.89 \\
Qwen 2.5‑14B‑Instruct    & 95.11 & 80.89 & 88.38 \\
Qwen 2.5‑32B‑Instruct    & 96.00 & 84.21 & 89.96 \\
Qwen 2.5‑72B‑Instruct    & 95.45 & 85.02 & 90.43 \\[2pt]
Phi‑3‑mini‑128k‑Instruct & 91.49 & 76.85 & 83.74 \\
Phi‑4‑mini‑Instruct      & 91.79 & 78.55 & 83.45 \\
Phi‑3‑small‑128k‑Instruct& 87.74 & 75.87 & 76.81 \\
Phi‑3‑medium‑128k‑Instruct& 94.85& 82.68 & 90.04 \\
\bottomrule
\end{tabular}
\end{table*}

\end{document}